%% file: main.tex
\PassOptionsToPackage{table}{xcolor}
\documentclass{adobe_research}
\input{math_commands.tex}

\usepackage{newtxtext}
\usepackage{url}
\usepackage{cuted}
\usepackage{makecell}
\usepackage[normalem]{ulem}
\definecolor{overallblue}{HTML}{246B50}

\title{MoviEdit: Long-form Multi-Shot Video Editing Made Easy}

\author{%
  Suhyeon Lee$^{1,2,*}$\quad
  Zhengang Li$^{1}$\quad
  Soo Ye Kim$^{1}$\quad
  Peihao Zhu$^{1}$\quad
  Jong Chul Ye$^{2}$\quad\\
  Chun-Hao Paul Huang$^{1}$\quad
  Duygu Ceylan$^{1}$\quad
  Yan Kang$^{1}$\quad
  Hyeonho Jeong$^{1,*,\dagger}$%
}

\affiliation{%
  $^{1}$Adobe\quad
  $^{2}$KAIST\\
  $^{*}$Equal Contribution\quad
  $^{\dagger}$Tech Lead%
}

\abstract{
\vspace{-2mm}
Long-form multi-shot video editing requires an intended edit to remain coherent across abrupt changes in scene, viewpoint, and subject appearance, while processing substantially longer visual context than conventional single-shot editors.
We introduce \textbf{MoviEdit}, an end-to-end framework for scalable long-form multi-shot video editing that jointly addresses the challenges of data, modeling, and evaluation. 
To provide cross-shot supervision data at scale, we develop automatic data-generation pipelines spanning synchronized multi-view footage, real-world multi-shot videos, and generated multi-shot content, yielding 87K paired editing videos without manual annotation. 
Architecturally, MoviEdit employs a cascaded bidirectional--autoregressive design: a lightweight bidirectional stage first establishes a globally coherent editing draft over a spatiotemporally compressed representation, which is then autoregressively propagated and refined over the full video. 
We further introduce \textbf{MoviEdit-Bench}, a dedicated benchmark for systematic evaluation of long-form multi-shot video editing.
Extensive experiments show that MoviEdit substantially outperforms existing open-source models and performs on par with industry frontier models across text-guided and image-guided editing. Visit our project page for complete results and to access the benchmark:
\textbf{\href{https://multi-shot-edit.github.io/}{multi-shot-edit.github.io}}
\vspace{-0.5mm}
}

\begin{document}
\maketitle

\begin{center}
\includegraphics[width=0.96\textwidth]{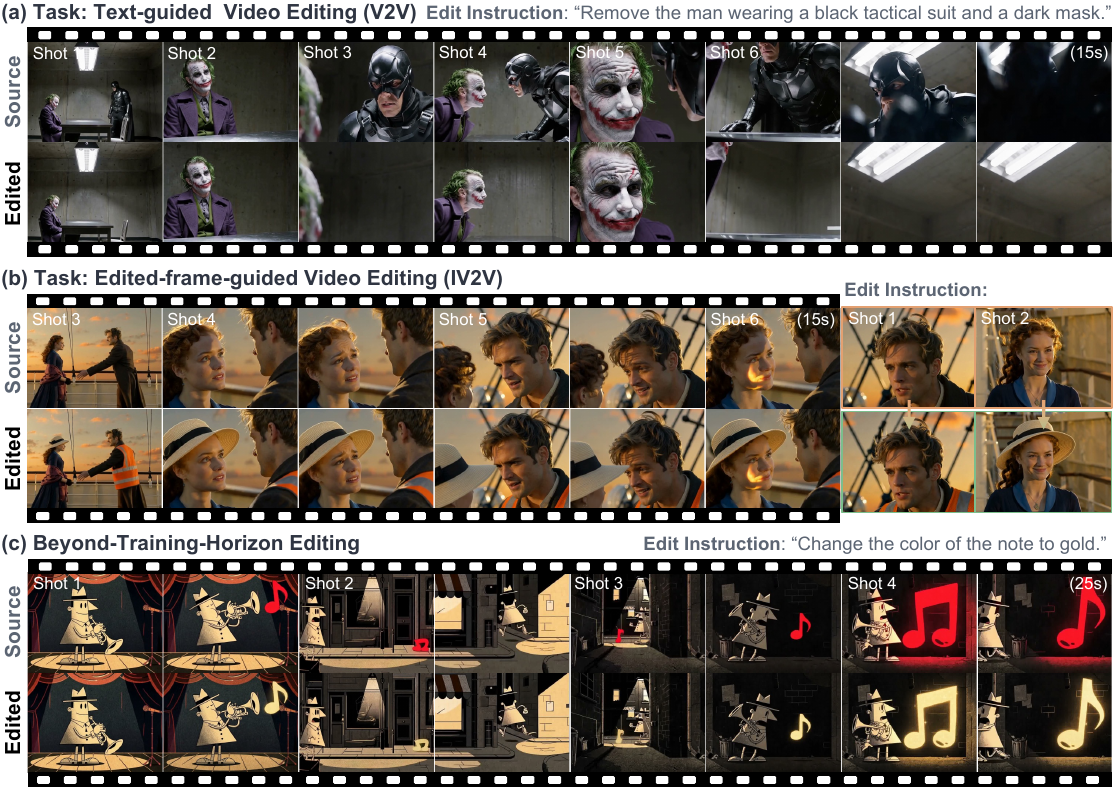}
\captionof{figure}{\textbf{Long-form multi-shot video editing with MoviEdit.}
MoviEdit supports (a) text-guided and (b) edited-frame-guided multi-shot video editing, enabling consistent edits across shots, and (c) generalizes to videos longer than those seen during training. 
}
\label{fig:title}
\end{center}

\section{Introduction}
Visual narratives are inherently compositional: their meaning emerges not only within individual shots, but also from the relationships among shots over time.
Recent advances in generative video modeling have begun to capture this structure, moving beyond short, isolated clips toward long-form, multi-shot video generation \citep{guo2025long, zhang2025storymem, cai2026mixture, meng2026holocine}.
Yet existing video editing methods \citep{decart2025lucyedit, jiang2025vace, lee2026memory, liu2025generative, ju2026editverse} remain predominantly designed for short, single-shot videos, typically spanning only a few seconds.
Extending editing to long-form, multi-shot content is not merely a matter of processing more frames: an intended edit must persist coherently across cuts even as the scene, viewpoint, motion, appearance, and surrounding context change.

The difficulty is structural rather than simply one of scale. 
Editing each shot independently discards the cross-shot dependencies needed to preserve a consistent edit, whereas jointly processing the full source and target sequences quickly becomes computationally prohibitive.
Yet this computational dilemma is only one part of the problem.
Fundamentally, long-form multi-shot editing exposes three coupled bottlenecks in existing video editing systems: \emph{data}, \emph{modeling}, and \emph{evaluation}.
At the \emph{data} level, learning to edit a subject consistently across shots requires source and target video pairs with explicit cross-shot subject correspondence. 
Such data does not exist in the real world and, to our knowledge, no existing pipeline can synthesize them at the scale and quality required for training \citep{zi2026senorita, bai2026scaling, wu2025insvie, lin2026kiwi, long2026coinve}.
On the \emph{modeling} side, long-form editing is fundamentally a memory problem. 
Computationally, editing typically requires processing a source-video representation alongside the target video being synthesized; under common token-concatenation designs, the quadratic cost of global self-attention rapidly becomes prohibitive as duration increases.
Third, on the \emph{evaluation} side, there is no dedicated evaluation protocol for this setting.
Existing video editing benchmarks~\citep{li2025five, bian2025videopainter, ju2026editverse} focus primarily on short, single-shot clips and therefore do not assess the capabilities that distinguish multi-shot editing, such as whether an edit continues to localize the correct subject after a cut and whether its appearance remains consistent when that subject reappears.
Once supervised data and evaluation are in place, the modeling problem becomes tractable with the goal of maintaining temporal and appearance consistency over long horizons.

To this end, we present \textbf{MoviEdit}, an end-to-end framework for training and evaluating long-form multi-shot video editing models.
MoviEdit addresses the three bottlenecks through a matched set of data, modeling, and evaluation components.
First, we develop a scalable data generation pipeline that combines multi-view-aware image editing, single-shot video editing, and VLM-based quality filtering to construct high-quality paired training examples with explicit subject correspondence across shots.
Second, we present a cascaded video editing architecture that combines bidirectional global context with causal propagation.
A lightweight bidirectional editing stage first constructs a globally coherent edit draft, after which a causal stage autoregressively propagates and refines the edit over the full sequence.
This decomposition avoids the prohibitive memory cost of jointly processing the complete source and target videos while preserving temporal and appearance consistency across shots.
Lastly, we introduce \textbf{MoviEdit-Bench}, a dedicated benchmark that spans multiple difficulty levels and evaluates both local, subject-specific edits and global, scene-level transformations. It provides automated metrics for cross-shot target localization and edit consistency, together with a curated collection of copyright-free multi-shot videos.
Together, these components provide the supervision, architecture, and evaluation infrastructure needed to move video editing beyond isolated clips toward coherent, long-form visual narratives.
Extensive experiments further demonstrate that a MoviEdit model trained using our framework substantially outperforms existing open-source solutions, and achieves performance on par with strong proprietary systems \citep{google2026geminiomni, alibaba2026happyhorse, runway2026aleph2} across both text-guided video editing and image-guided video editing.

Our contributions are summarized as follows:
\begin{itemize}
    \setlength{\itemsep}{1pt}
    \setlength{\parskip}{0pt}
    \setlength{\parsep}{0pt}
    \item We introduce a scalable paired-data generation pipeline for supervised long-form multi-shot video editing. 

    \item We propose an efficient long-form video editing architecture based on a cascaded bidirectional--autoregressive diffusion framework. 

    \item We present MoviEdit-Bench, a dedicated benchmark and evaluation suite for long-form multi-shot video editing. 
\end{itemize}

\section{Related Work} \label{sec:supp-related-work}
\textbf{Diffusion-based Video Editing.}
Early diffusion-based video editing methods largely repurpose pretrained generative models as editing priors through inference-time adaptation. Typical strategies include inversion \citep{qi2023fatezero, wang2023zero}, attention or feature manipulation \citep{qi2023fatezero, liu2024video, geyer2024tokenflow, ceylan2023pix2video, jeong2024dreammotion}, auxiliary visual conditioning \citep{zhang2023adding, jeong2023ground}, and per-video optimization \citep{wu2023tune, jeong2024vmc, jeong2025reangle}, enabling edits without requiring large-scale paired training data. While effective in many settings, these methods rely primarily on manipulating pretrained generative priors at inference time rather than learning generalizable editing behavior from data, which can limit robustness, scalability, and editing flexibility while introducing substantial inference overhead.

More recently, video editing has shifted toward a data-driven paradigm, where models learn editing behavior from paired source--target videos \citep{decart2025lucyedit, ju2026editverse, jiang2025vace, liu2025generative, polyak2024movie, yang2026omni, seo2026propfly, lin2026kiwi, lee2026memory, huang2025spacetimepilot}.
These methods typically build on pretrained video generators and differ mainly in how source-video information is incorporated, ranging from channel-level \citep{decart2025lucyedit} or token-level concatenation \citep{ju2026editverse, lee2026memory} to adapter-based additive conditioning~\citep{jiang2025vace}.
They also vary in how the edit is specified; for example, GenProp~\citep{liu2025generative} conditions on an edited reference frame and propagates the corresponding transformation through the video.
Following the same data-driven philosophy, our models use token-level source conditioning and are trained separately for text-guided and edited-image-guided editing.
Unlike prior reference-image-guided methods \citep{ouyang2024i2vedit, ku2024anyv2v, liu2025generative} that typically condition on an edited first frame, our image-guided model supports edited references from arbitrary frames. 
Complementary approaches introduce explicit reasoning or agentic control into the editing process, from latent reasoning over edit regions~\citep{yang2026videocof} to LLM/VLM-driven shot-level planning for multi-shot editing~\citep{wu2026thinking}.

More broadly, while most existing end-to-end video editing models remain focused on short, single-shot settings, our work systematically addresses long-form multi-shot editing through cross-shot supervision, long-context modeling, and dedicated evaluation.

\textbf{Long-Form and Multi-Shot Video Synthesis.}
Recent advances have made long-form video generation increasingly tractable through autoregressive generation and efficient long-context modeling. 
Methods such as CausVid~\citep{yin2025slow} and Self-Forcing~\citep{huang2026self} generate videos autoregressively in chunks, using windowed attention or rolling KV caches to bound the context and computational cost over long horizons.
In parallel, SANA-Video~\citep{chen2026sana} and SANA-Streaming~\citep{zhao2026sana} replace softmax attention with linear attention~\citep{katharopoulos2020transformers}, enabling efficient long-context modeling and constant-memory, constant-compute streaming generation. 
Few-step diffusion distillation has also been explored to reduce autoregressive sampling latency, including DMD- and consistency-based methods~\citep{huang2026self, zhao2026causal} and, more recently, velocity moment matching (VMM)~\citep{zhou2026editstream}.

Beyond a continuous sequence, recent work has begun to address multi-shot video generation by extending conventional short-form video models with two key modifications: training on multi-shot video data and introducing shot-aware sparse attention for efficient cross-shot context modeling. HoloCine~\citep{meng2026holocine} employs dense intra-shot attention while using the first-frame tokens of each shot as compact summary tokens for cross-shot information exchange. Mixture-of-Contexts~\citep{cai2026mixture} further introduces content-adaptive sparse attention that dynamically retrieves a small set of relevant historical chunks for each query, while ContextMaster~\citep{guo2026contextmaster} adopts fixed-budget context routing that combines mandatory task-specific context with dynamically selected historical information. MSEditor~\citep{feng2026mseditor} instead leverages synchronized multi-view data to adapt a first-frame-conditioned video model for consistent appearance propagation across multiple shots. 

Despite their different implementations, these approaches improve scalability by limiting dense global interactions across the video, making it difficult to maintain a consistent appearance over long temporal spans. This highlights the need for modeling strategies that balance the computational efficiency of autoregressive generation with the global context provided by bidirectional modeling.

\section{MoviEdit: Long-Form Multi-Shot Video Editing}
\begin{figure}[!t]
    \centering
    \includegraphics[width=1.0\textwidth]{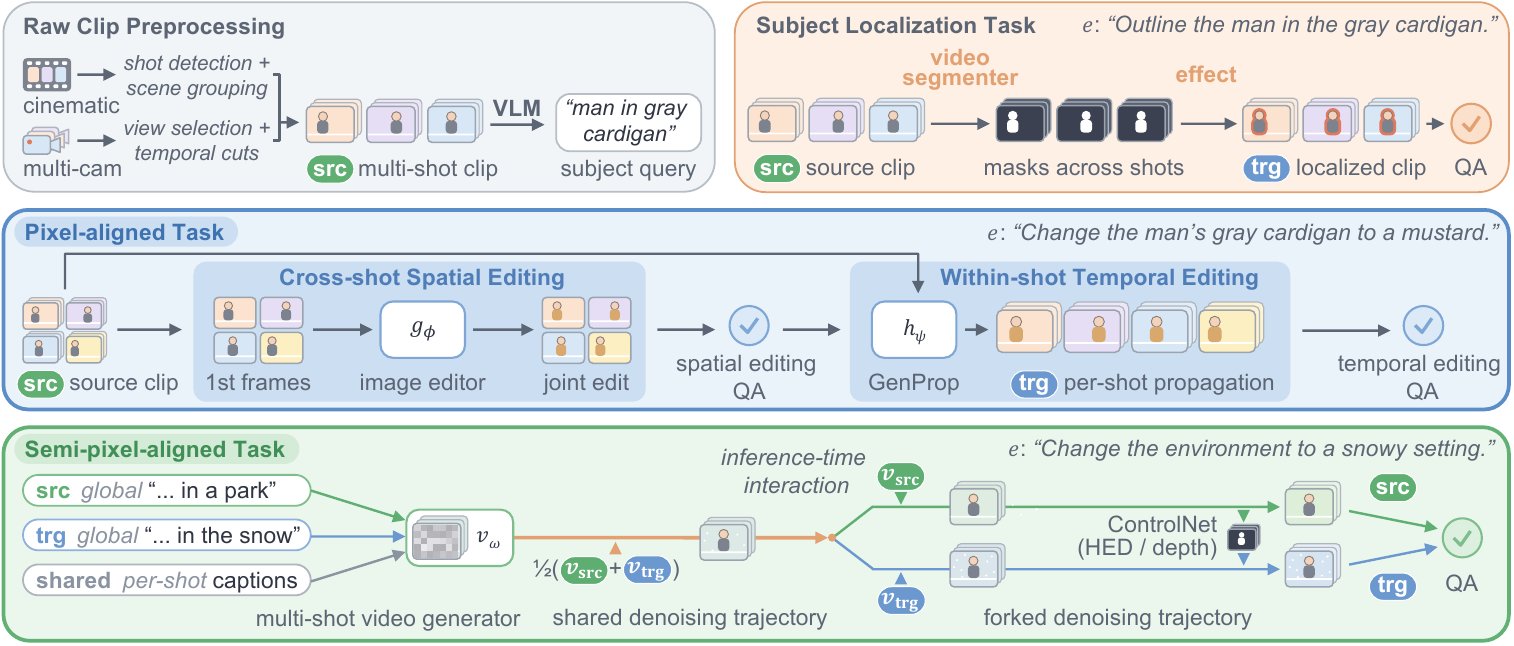}
    \caption{\textbf{Scalable paired-data generation for multi-shot video editing.}
Three annotation-free pipelines transform diverse video sources into large-scale paired data for consistent multi-shot editing. Pixel-aligned editing uses a spatial--temporal editing decomposition to ensure cross-shot consistency, while semi-pixel-aligned editing uses coupled multi-shot video generation with inference-time interaction to produce structurally consistent source--target pairs.}
\label{fig:method-data}
\end{figure}

\begin{figure}[!t]
    \centering
    \includegraphics[width=1.0\textwidth]{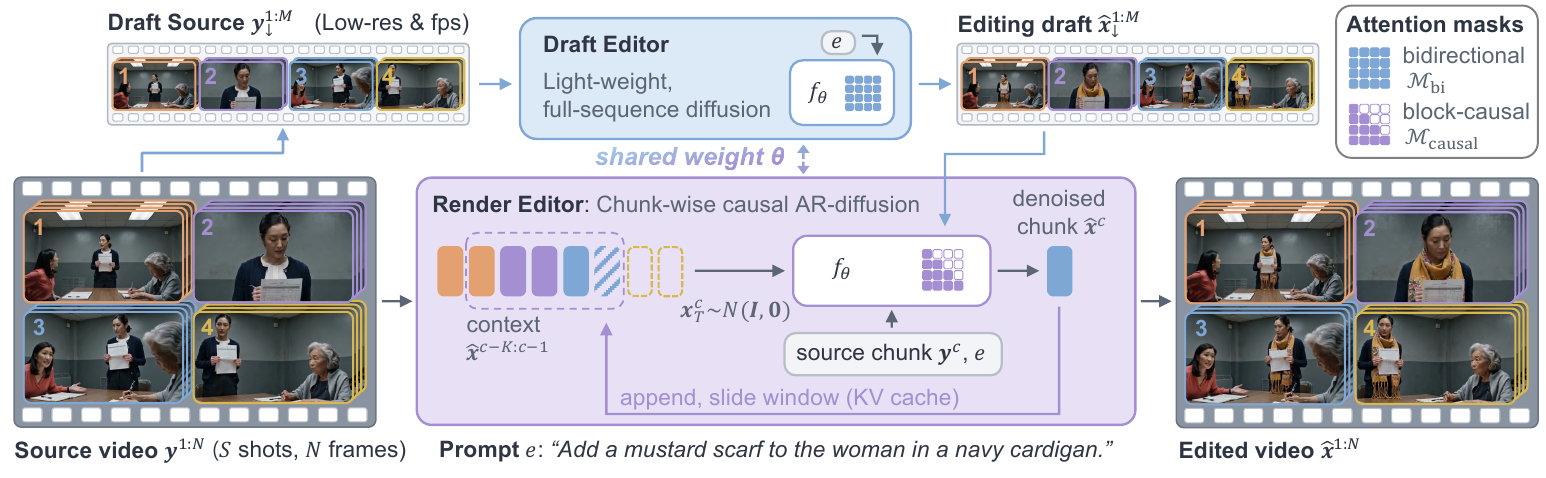}
    \caption{\textbf{Cascaded modeling for a long-form video editor.} 
    MoviEdit uses a shared denoiser $f_\theta$ to first produce a globally consistent, lightweight edit on a subsampled video sequence (Draft Editor), and then propagates the edit chunk by chunk to the full-resolution video while mitigating long-horizon appearance drift (Render Editor).}
    \label{fig:method-model}
\end{figure}

MoviEdit provides an end-to-end framework that develops the key ingredients for building industry-level long-form multi-shot video editors.
We first describe our scalable paired-data generation pipelines for multi-shot video editing tasks (Fig.~\ref{fig:method-data}), followed by the cascaded architecture for a long-form video editing model (Fig.~\ref{fig:method-model}).
We further describe several complementary techniques for improving inference efficiency.
Finally, we introduce MoviEdit-Bench for systematic evaluation of cross-shot editing capabilities. 

\subsection{Scalable Paired Data Generation for Multi-Shot Video Editing}

\begin{figure}
    \centering
    \vspace{-1.5em}
\includegraphics[width=0.9\textwidth]{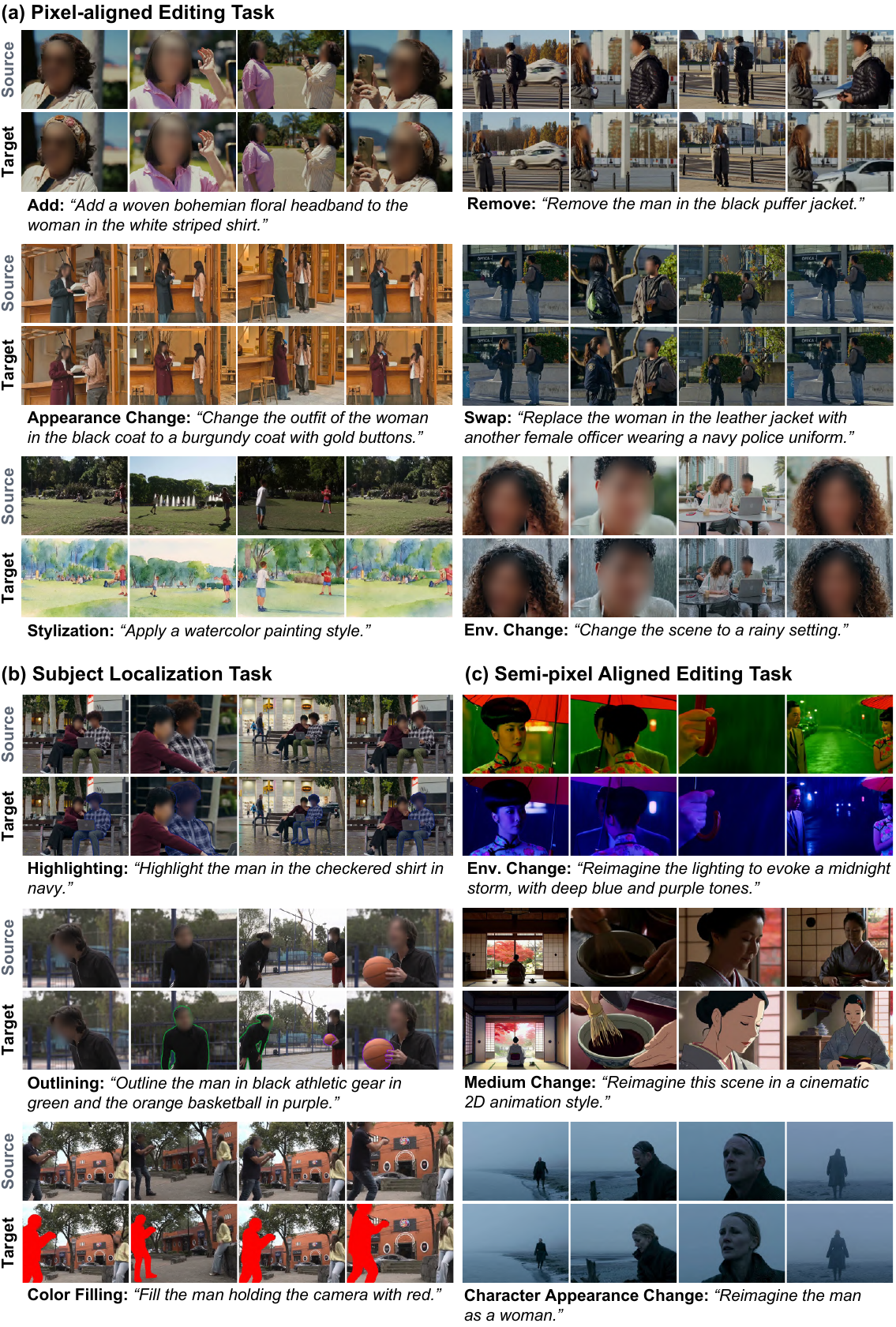}
\caption{
\textbf{Representative training data pairs across editing tasks.} We show source and target video examples from our three data-generation pipelines. The pixel-aligned pipeline covers add, remove, appearance change, swap, stylization, and environment change; the localization pipeline covers highlighting, outlining, and color filling; and the semi-pixel-aligned pipeline covers environment change, medium change, and character appearance change. \textit{Note: Faces of real actors in the displayed examples are blurred to protect their identity.}
}
\label{fig:supp-data}
\end{figure}

We develop three scalable pipelines for constructing paired supervision without human annotation (Figs.~\ref{fig:method-data} and~\ref{fig:supp-data}).
We leverage in-house synchronized multi-camera 
footage (Fig.~\ref{fig:supp-data-source}),
real-world cinematic clips, or pretrained multi-shot text-to-video diffusion models. 
These sources offer complementary strengths in cross-view correspondence, cinematic realism, and controllable synthesis, respectively, supporting three supervision regimes: \textit{pixel-aligned editing}, \textit{subject localization}, and \textit{semi-pixel-aligned editing}. Detailed training-data statistics are provided in Appx.~\ref{sec:supp-data-source}, with data-recipe ablations in Fig.~\ref{fig:data_ablation}.

For pixel-aligned editing and subject localization pipelines, we obtain single-scene, multi-shot clips either from real-world cinematic clips using shot detection~\citep{wang2026omnishotcut} and scene grouping~\citep{mun2022bassl}, or from synchronized multi-camera footage through camera-view selection and temporal cuts.
A vision-language model (VLM)~\citep{yang2025qwen3, team2026gemma} proposes candidate subjects as short text queries, merging observations of the same entity across shots into one query. The multi-shot clips and subject queries are used to construct supervision for both tasks.

\textbf{Pixel-aligned Editing.}
The pixel-aligned editing task is our primary setting, where the edited video preserves pixel-wise spatial correspondence with the source while modifying only the instructed visual content. For object-level edits, including \textit{add}, \textit{remove}, \textit{swap}, and \textit{appearance change}, we use the subject query to select the target instance to edit in the first frame of each shot. For global edits, including \textit{environment change} and \textit{stylization}, we apply the edit to the full first frame of each shot.

Ensuring cross-shot consistency is a critical challenge in this setting. We address it by decomposing multi-shot editing into two complementary subproblems: \textit{cross-shot spatial editing}, which resolves a consistent edit across representative frames from all shots, and \textit{within-shot temporal editing}, which propagates each resolved edit through its corresponding shot. For the first stage, we arrange the first frames $\bm{y}_{s}^{1}$ from all $S$ source video shots $\{\bm{y}_{s}\}_{s=1}^{S}$ into a single image grid $[\bm{y}_{1}^{1}; \dots; \bm{y}_{S}^{1}]$ and jointly edit them in a single inference pass using off-the-shelf text-guided image editing models $g_{\phi}$. Each edited first frame is then propagated independently within its corresponding shot using a pretrained first-frame-conditioned single-shot video editing model $h_{\psi}$:
\begin{equation}
\label{eq:pixel_aligned}
\tilde{\bm{y}}_{1:S}^{1}
=
[\tilde{\bm{y}}_{1}^{1};\dots;\tilde{\bm{y}}_{S}^{1}]
=
g_\phi\!\left(
[\bm{y}_{1}^{1};\dots;\bm{y}_{S}^{1}], e
\right),
\quad
\bm{x}_{s}
=
h_\psi\!\left(
\tilde{\bm{y}}_{s}^{1}, \bm{y}_{s}, e
\right),
\quad s=1,\dots,S,
\end{equation}
where $e$ denotes the edit instruction, and $\bm{x}_s$ is the target edited video shot. 
This spatial--temporal decomposition avoids bootstrapping from an existing multi-shot video editor.
We apply quality filtering at both stages: first-frame QA checks edit success and cross-shot consistency, while propagation QA checks within-shot consistency and overall video consistency after recomposition, rejecting $17\%/38\%$ of multi-camera and $32\%/47\%$ of cinematic pairs at the two stages, respectively.

\textbf{Subject Localization.}
This task, also supported by recent video editing models ~\citep{ju2026editverse, liu2025generative}, requires identifying a queried subject across multiple shots and applying localized visual transformations, including effects such as \textit{highlighting}, \textit{outlining}, and \textit{color filling}.
Given a multi-shot source video and subject query, video segmentation models ~\citep{cheng2024putting, sari2025flowcut, carion2026sam} track the subject across all shots where it appears and produce dense per-frame masks.
Applying the specified effect to these masks yields paired source--target videos for supervising cross-shot localization.
With VLMs, we further filter the generated pairs to ensure consistent subject localization and correct effect application, rejecting $24\%$ of pairs.

\textbf{Semi-pixel-aligned Editing.}
This pipeline generates paired source--target videos that preserve spatial structure, temporal composition, and narrative correspondence without requiring strict pixel-level alignment.
By relaxing exact correspondence, it supports a broader range of edits while increasing data diversity through coupled generation with a pretrained text-driven multi-shot video generator (e.g.,~\citealp{meng2026holocine, zhang2025storymem}).
We apply this pipeline to tasks including \textit{environment change}, \textit{medium change}, and \textit{character appearance change}.

For each pair, we couple source and target generation by sharing the same per-shot captions and initial noise while using distinct global captions. An LLM constructs the source and target global captions by modifying only the attribute associated with the intended edit while preserving all other content. To further couple them, we introduce task-specific inference-time interaction mechanisms. 
For \textit{character appearance change}, the source and target instead share a common denoising trajectory until a fork timestep $t_{\mathrm{fork}}$. For $t > t_{\mathrm{fork}}$, the shared latent is updated using the averaged source- and target-conditioned velocity predictions,
\begin{equation}
\label{eq:branching}
\tilde{\bm{v}}(\bm{z}_t,t)
=
\left(
\bm{v}_\omega(\bm{z}_t,e_{\mathrm{src}},t)
+
\bm{v}_\omega(\bm{z}_t,e_{\mathrm{tgt}},t)
\right)/2.
\end{equation}
For $t \leq t_{\mathrm{fork}}$, the two generations branch from the shared latent state and are denoised independently under their respective conditions. 
For environment change and medium change editing, the target generation is additionally conditioned via ControlNet~\citep{zhang2023adding} on either an HED edge or a depth map extracted from the source video. 
These guarantee necessary structural correspondence while allowing the target appearance to diverge.

We further apply VLM-based quality control to generated pairs, filtering them based on prompt alignment, video quality, cross-shot consistency, and source--target distinctiveness, rejecting $66\%$ of pairs. For ControlNet-based pairs, we swap source and target when forming training pairs, using the structurally conditioned generation as input and the original generation as the reconstruction target, since the latter receives higher visual-quality scores.

\begin{figure}[!t]
    \centering
    \includegraphics[width=1.0\textwidth]{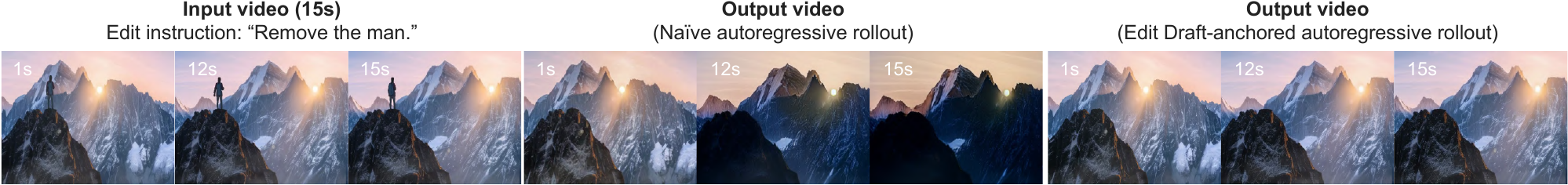}
    \caption{
    \textbf{Long-horizon appearance-drift stress test.}
    We compare autoregressive editing with and without draft conditioning in a small-scale train setting.
    Naive autoregressive editing accumulates appearance drift and saturation, while draft-conditioned rollout maintains a stable edited appearance.
    }
    \label{fig:drift-comparison}
\end{figure}

\subsection{Cascaded Architecture for Long-Form Video Editing}

A common design for long-form video modeling is chunk-wise autoregressive diffusion~\citep{yin2025slow, liu2026rolling}, where a video is divided into $C$ fixed-size chunks $\bm{x}^{1:C}$ and modeled with a fixed-size sliding context window over recent chunks. In causal diffusion forcing (CDF)~\citep{chen2024diffusion}, each video chunk $\bm{x}^{c}$ is independently corrupted at a sampled diffusion timestep $t_c$, yielding
$
\bm{x}_{t_c}^{c}
=
\alpha_{t_c}\bm{x}^{c}
+
\sigma_{t_c}\bm{\epsilon}^{c},
$
and the denoiser is trained with
\begin{equation}
\label{eq:cdf}
\mathcal{L}_{\mathrm{CDF}}(\theta)
=
\mathbb{E}_{\bm{x},\bm{t},\bm{\epsilon}}
\left[
\sum_{c=1}^{C}
\left\|
\left[
f_\theta\!\left(
\bm{x}_{\bm{t}}^{1:C};
\mathcal{M}_{\mathrm{causal}}
\right)
\right]^{c}
-
\bm{v}_{t_c}^{c}
\right\|_2^2
\right],
\qquad
t_c \overset{\mathrm{i.i.d.}}{\sim} p(t),
\end{equation}
where
$\bm{t}=(t_1,\dots,t_C)$,
$\bm{x}_{\bm{t}}^{1:C}
=
\{\bm{x}_{t_c}^{c}\}_{c=1}^{C}$
and $\bm{v}_{t_c}^{c}$ denotes the diffusion target (\textit{e.g.}, velocity) for chunk $c$ at timestep $t_c$.
The causal self-attention mask $\mathcal{M}_{\mathrm{causal}}$ ensures that a standard DiT-based denoiser~\citep{peebles2023scalable} $f_\theta$ predicts each chunk $c$ using only itself and the preceding chunks within the retained context window.
During inference, the same denoiser $f_\theta$ is used within a chunk-wise causal autoregressive diffusion sampler,
\begin{equation}
\label{eq:sampling}
\hat{\bm{x}}^{c}
=
\mathcal{S}_{f_\theta}\!\left(
\left[
\hat{\bm{x}}^{c-K:c-1};
\bm{x}^{c}_{T}
\right];
\mathcal{M}_{\mathrm{causal}}
\right),
\qquad
\bm{x}^{c}_{T}\sim\mathcal{N}(\mathbf{0},\mathbf{I}),
\end{equation}
where $\mathcal{S}_{f_\theta}$ denotes a standard diffusion sampler using $f_\theta$, $[\cdot\,;\,\cdot]$ denotes temporal token concatenation, and $\hat{\bm{x}}^{c-K:c-1}$ denotes the preceding $K$ generated chunks retained as autoregressive context.

However, the limited global context of chunk-wise autoregressive generation leads to cumulative appearance drift and oversaturation over long editing horizons (Fig.~\ref{fig:drift-comparison}).
To mitigate this issue, we augment the efficient CDF backbone with a lightweight bidirectional diffusion stage that provides global guidance before full-resolution autoregressive editing.
The \textit{Draft Editor} first performs full-sequence bidirectional diffusion over a spatially and temporally subsampled video representation t, producing a globally coherent edit draft. 
This draft serves two purposes: it provides stable appearance anchors that suppress long-horizon drift during autoregressive generation, and, by jointly resolving the edit over the entire sequence, establishes cross-shot consistency even between temporally distant shots.
The \textit{Render Editor} then performs full-resolution chunk-wise causal diffusion conditioned on the draft, preserving this global editing structure while efficiently synthesizing local detail from the full-resolution input video chunks (Fig.~\ref{fig:method-model}).
The \textit{Draft Editor} and \textit{Render Editor} share model weights, differing only in their attention patterns and spatiotemporal resolutions.

Let $\bm{y}^{1:N}$ denote the $N$-frame source video and $e$ the edit instruction. 
We apply a fixed spatiotemporal subsampling operator $\downarrow$ to obtain a shorter, lower-resolution video sequence $\bm{y}_{\downarrow}^{1:M} = {\downarrow}(\bm{y}^{1:N})$, where $M{\ll}N$.
Empirically, we find $M=N/2$ to be effective.

\textbf{Draft Editing.} The Draft Editor applies $f_\theta$ bidirectionally to the full subsampled sequence:
\begin{equation}
\label{eq:stage1}
\hat{\bm{x}}_{\downarrow}^{1:M}
=
\mathcal{S}_{f_\theta}\!\left(
\bm{x}^{1:M}_T;
\bm{y}_{\downarrow}^{1:M},
e;
\mathcal{M}_{\mathrm{bi}}
\right),
\qquad
\bm{x}^{1:M}_{T}\sim\mathcal{N}(\mathbf{0},\mathbf{I}),
\end{equation}
where $\mathcal{M}_{\mathrm{bi}}=\mathbf{0}$ removes causal masking, allowing full bidirectional attention across the subsampled sequence. 
The resulting draft jointly resolves the edit over the entire video, providing a global appearance anchor for subsequent rendering.

\textbf{Render Editing.} 
Conditioned on this draft edit, the Render Editor applies $f_\theta$ causally over fixed-size chunks $c=1,\dots,C$ of the full-resolution target video,
\begin{equation}
\label{eq:stage2}
\hat{\bm{x}}^{c}
=
\mathcal{S}_{f_\theta}\!\left(
[\hat{\bm{x}}^{c-K:c-1};\bm{x}_T^c];
\hat{\bm{x}}_{\downarrow}^{1:M},
\bm{y}^{c},
e;
\mathcal{M}_{\mathrm{causal}}
\right),
\qquad
\bm{x}^{c}_{T}\sim\mathcal{N}(\mathbf{0},\mathbf{I}),
\end{equation}
where $\hat{\bm{x}}^{c-K:c-1}$ denotes the preceding $K$ generated chunks retained as autoregressive context. 
Each chunk conditions on the recently generated context for temporal continuity, the global draft for long-range appearance and cross-shot consistency, and the corresponding source chunk $\bm{y}^{c}$ to preserve content outside the intended edit.

During training, Draft Editor uses a conventional flow-matching objective \citep{lipman2022flow} with full bidirectional attention, while Render Editor is optimized with the CDF objective in Eq.~\ref{eq:cdf}.

\subsection{Inference Optimization}
Our architecture supports complementary techniques for reducing inference latency, including few-step diffusion distillation and KV caching, where 
efficiency gains are reported in Tab.~\ref{tab:efficiency_ablation}.

\textbf{Unified Distillation with Asymmetric Step Schedules.}
After teacher training, we follow velocity moment matching (VMM)~\citep{zhou2026editstream, NEURIPS2024_3f66d5cd} to jointly distill the bidirectional Draft Editor and causal Render Editor under the shared-weight architecture. 
Following VMM, timestep pairs are sampled according to a predefined student inference schedule; we assign a separate schedule to each editing stage within the same distillation run.
Specifically, the Draft Editor is distilled to $12$ denoising steps, while the more computationally expensive Render Editor is distilled to $4$ steps. 
We empirically find that the Render Editor is more robust to aggressive step reduction.
This asymmetric allocation preserves sufficient capacity for establishing global edit consistency while aggressively reducing the dominant rendering cost.

\textbf{KV Caching.}
During Render Editor inference, the autoregressive formulation and causal attention masking naturally support KV caching. 
We cache key-value states for all reusable tokens, including text, edit draft, source video, and previously generated chunk tokens, eliminating recomputation.

\begin{figure}[t]
    \centering
    \includegraphics[width=0.9\textwidth]{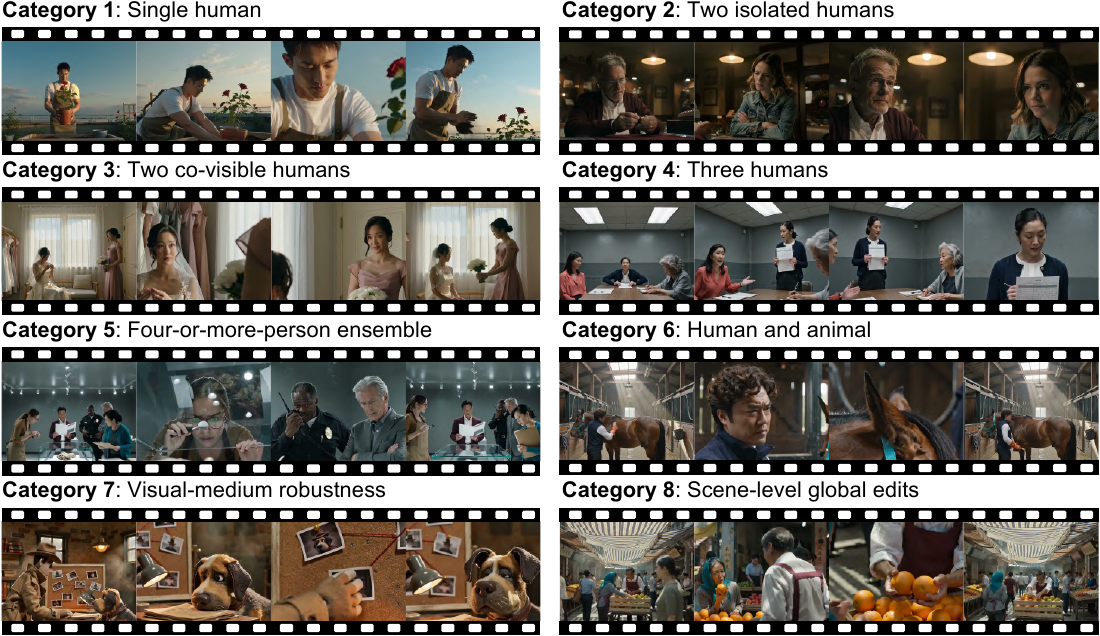}
    \caption{
    \textbf{Example source videos from MoviEdit-Bench.}
    The examples highlight the diversity of subject composition, cross-shot grounding difficulty, and visual appearance in the benchmark. Full videos are available on our project page.
    }
    \label{fig:supp-bench-source-cat}
\end{figure}

\subsection{MoviEdit-Bench: Evaluating Long-Form Multi-Shot Video Editing}

Existing benchmarks for instruction-guided video editing~\citep{bian2025videopainter, li2025five, ju2026editverse} primarily focus on short, single-shot clips and therefore do not explicitly evaluate cross-shot subject localization and editing consistency over long multi-shot sequences. To address this gap, we introduce \textbf{MoviEdit-Bench}, a benchmark for long-form cinematic multi-shot video editing under practical production settings, intended for public release. Benchmark construction and evaluation details are provided in Appx.~\ref{sec:supp-bench}.

MoviEdit-Bench contains 64 generated multi-shot videos and 128 editing scenarios spanning cinematic and drama-style content. The videos are organized into eight categories with varying cross-shot subject-binding difficulty, ranging from a single recurring subject to crowded multi-subject and mixed human--animal scenes, alongside visual-medium robustness and scene-level global-edit cases (Fig.~\ref{fig:supp-bench-source-cat}). The benchmark supports both text-guided video editing (V2V) and arbitrary-frame edited-frame-guided propagation (IV2V), with first-frame and non-first-frame reference scenarios. It covers both local edits (e.g., \textit{add}, \textit{remove}, \textit{appearance change}, and \textit{swap}) and global edits (e.g., \textit{environment change} and \textit{stylization}).

\textbf{Evaluation Protocol.}
To capture the diverse failure modes of long-form multi-shot editing, our evaluation spans seven complementary dimensions: \textit{non-subject preservation}, \textit{edit localization}, \textit{subject preservation}, \textit{cross-shot edit consistency}, \textit{edit-prompt alignment}, \textit{video quality}, and, for IV2V only, \textit{edited-frame fidelity}.
Across these dimensions, we combine pixel-level metrics, embedding-based similarity, and judgments from a vision-language model (VLM)~\citep{team2026gemma, yang2025qwen3}, following prior video editing benchmarks~\citep{ju2026editverse, li2025five, bian2025videopainter}. To account for multi-shot structure, source and edited videos are spatially and temporally aligned, with temporal metrics computed independently within each shot. The applicability of individual metrics varies by edit type, as detailed in Appx.~\ref{sec:supp-metrics}.

For frame-pair VLM evaluation, the judge examines uniformly sampled source and edited frame pairs from each shot, together with the edit instruction and target description, to produce fine-grained binary judgments on subject visibility, edit presence, correct subject localization, edit coverage, and preservation of unrelated content and non-edited subject attributes. Frames from shots where the subject is absent are also evaluated to detect unintended edits to other subjects or shots. For global edits, the corresponding judgments assess scene-level edit presence, coverage, and preservation of the underlying scene content. These criterion-level judgments are aggregated into VLM-based metrics for multiple evaluation dimensions, while cross-shot edit consistency is assessed separately by jointly evaluating edited frames across shots at the scenario level.

\textbf{Evaluation Metrics.} \textit{Non-subject preservation} measures unintended changes outside the intended edit region using LPIPS~\citep{zhang2018unreasonable} between source and edited frames, excluding the subject region with segmentation masks. \textit{Edit localization} evaluates whether modifications are confined to the intended subject and sufficiently cover it, using VLM-based Precision and Recall, respectively. Precision measures preservation of unrelated content, including competing subjects and subject-absent shots, whereas Recall measures completeness of the edit on the subject. For global edits, Recall instead measures coverage of the entire scene.

\textit{Subject preservation} assesses whether subject properties unrelated to the requested edit remain intact. We measure motion preservation using the Motion Fidelity Score (MFS)~\citep{yatim2024space}, which compares tracked point displacements between source and edited videos independently within each shot. For applicable local edits, LPIPS measures preservation of the original subject appearance, while VLM judgments assess preservation of non-edited subject attributes. \textit{Cross-shot edit consistency} evaluates whether the same edit is maintained across shots. The VLM judge jointly examines edited frames sampled from all relevant shots to assess consistency of the resulting appearance, object, identity, or scene transformation, independently of edit correctness.

\textit{Edit-prompt alignment} measures correspondence between the edited video and the intended editing outcome using CLIP~\citep{radford2021learning} and ViCLIP~\citep{wang2024internvid} similarities against per-shot target-video descriptions, together with VLM judgments of edit presence. \textit{Video quality} is assessed independently of edit correctness through DINOv2~\citep{oquab2023dinov2} feature similarity between adjacent frames within each shot, NIQE~\citep{mittal2012making}, PickScore~\citep{kirstain2023pick}, and VLM-based visual quality assessment. Finally, \textit{edited-frame fidelity}, applicable only to IV2V, measures whether the propagated edit preserves the identity and attributes established by the edited reference frame while allowing natural variations in viewpoint, pose, lighting, and motion.

\textit{Overall edit success} summarizes editing performance by requiring each sampled frame to satisfy all applicable VLM criteria for edit presence, correct subject localization, edit coverage, preservation of unrelated content and non-edited subject attributes, and fidelity to the edited reference frame (IV2V only). Let $\mathcal{F}$ denote the set of sampled frames for a given editing scenario and $\mathcal{C}(t)$ the set of applicable VLM criteria for frame $t$. We define
{
\setlength{\belowdisplayskip}{1pt}
\setlength{\belowdisplayshortskip}{1pt}
\begin{equation}
\label{eq:vlmsuccess}
\texttt{overall\_edit\_success}
=
\frac{1}{|\mathcal{F}|}
\sum_{t \in \mathcal{F}}
\mathbf{1}\!\left[
\bigwedge_{c \in \mathcal{C}(t)}
a_c^t
\right],
\end{equation}
}
where $a_c^t$ denotes the VLM judge's binary verdict for criterion $c$ at sampled frame $t$. The score is computed using the VLM judgments from the individual evaluation dimensions, without additional queries.

\begin{table*}[t]
    \centering
    \setlength{\tabcolsep}{0pt} %
    \renewcommand{\arraystretch}{1.05}
    \scriptsize
    \caption{\textbf{Quantitative evaluation on the long-form multi-shot video editing benchmark, MoviEdit-Bench, for text-guided video editing (V2V).} (X-shot: cross-shot)}
    \label{tab:movie_edit_bench_v2v}
    \resizebox{0.98\textwidth}{!}{%
    \begin{NiceTabular}{@{}>{\centering\arraybackslash}p{0.5cm}l*{15}{c}@{}}
    \CodeBefore
    \columncolor{overallblue!22}{17}
    \Body
        \toprule
        \multirow{2}{*}{} & \multirow{2}{*}{\textbf{Model}} & \makecell{\textbf{Non-subject}\\\textbf{Preservation}} & \multicolumn{2}{c}{\makecell{\textbf{Edit}\\\textbf{Localization}}} & \multicolumn{3}{c}{\makecell{\textbf{Subject}\\\textbf{Preservation}}} & \makecell{\textbf{X-shot Edit}\\\textbf{Consistency}} & \multicolumn{3}{c}{\makecell{\textbf{Edit Prompt}\\\textbf{Alignment}}} & \multicolumn{4}{c}{\makecell{\textbf{Video}\\\textbf{Quality}}} & \makecell{\textbf{Overall Edit}\\\textbf{Success}} \\
        \cmidrule(lr){3-3} \cmidrule(lr){4-5} \cmidrule(lr){6-8} \cmidrule(lr){9-9} \cmidrule(lr){10-12} \cmidrule(lr){13-16} \cmidrule(lr){17-17}
        & & LPIPS$\downarrow$ & Precision$\uparrow$ & Recall$\uparrow$ & MFS$\uparrow$ & LPIPS$\downarrow$ & VLM$\uparrow$ & VLM$\uparrow$ & CLIP$\uparrow$ & ViCLIP$\uparrow$ & VLM$\uparrow$ & DINO$\uparrow$ & NIQE$\downarrow$ & PickS.$\uparrow$ & VLM$\uparrow$ & VLM$\uparrow$ \\
        \midrule
        \multirow{5}{*}{\rotatebox[origin=c]{90}{Open source}} & SDEdit (HoloCine) & 0.311 & 0.321 & 0.409 & 0.535 & 0.364 & 0.315 & 0.894 & 0.222 & 0.184 & 0.491 & 0.988 & 5.56 & 19.4 & 0.711 & 0.193 \\
        & Memory-V2V & 0.179 & 0.525 & 0.252 & 0.630 & 0.199 & 0.625 & 0.452 & 0.215 & 0.174 & 0.437 & 0.985 & \textbf{4.82} & 19.2 & 0.289 & 0.215 \\
        & Lucy Edit & 0.174 & 0.729 & 0.192 & 0.647 & 0.185 & 0.751 & 0.750 & 0.213 & 0.171 & 0.351 & 0.984 & 5.82 & 19.2 & 0.508 & 0.235 \\
        & SANA-Streaming & 0.141 & 0.651 & 0.430 & 0.651 & 0.212 & 0.707 & 0.490 & 0.219 & 0.181 & 0.592 & 0.984 & 5.48 & 19.3 & 0.504 & 0.354 \\
        & Wan-Edit & \underline{0.0816} & 0.840 & 0.355 & \underline{0.686} & \textbf{0.103} & 0.809 & 0.692 & 0.217 & 0.178 & 0.515 & 0.983 & 5.38 & 19.3 & 0.568 & 0.378 \\
        \midrule
        \multirow{3}{*}{\rotatebox[origin=c]{90}{Private}} & Gemini Omni Flash & 0.153 & 0.705 & \underline{0.834} & 0.568 & 0.193 & 0.684 & 0.893 & 0.221 & \textbf{0.194} & 0.834 & 0.987 & 4.95 & \textbf{19.8} & \textbf{0.814} & 0.586 \\
        & HappyHorse-1.0 & 0.246 & 0.729 & 0.788 & 0.400 & 0.291 & 0.728 & 0.894 & 0.219 & \underline{0.193} & 0.815 & 0.984 & 5.12 & 19.6 & 0.691 & 0.623 \\
        & Aleph 2.0 & 0.0916 & \underline{0.886} & 0.778 & 0.671 & 0.146 & \underline{0.883} & \textbf{0.981} & \underline{0.225} & 0.186 & \underline{0.846} & \underline{0.988} & 4.95 & 19.6 & \underline{0.801} & \underline{0.764} \\
        \midrule
         & \textbf{MoviEdit (Ours)} & \textbf{0.0741} & \textbf{0.904} & \textbf{0.861} & \textbf{0.695} & \underline{0.121} & \textbf{0.911} & \underline{0.904} & \textbf{0.227} & 0.190 & \textbf{0.884} & \textbf{0.989} & \underline{4.85} & \underline{19.7} & 0.762 & \textbf{0.790} \\
        \bottomrule
    \end{NiceTabular}%
    }
    \vspace{1em}

    \centering
    \setlength{\tabcolsep}{0pt} %
    \renewcommand{\arraystretch}{1.05}
    \scriptsize
    \caption{\textbf{Quantitative evaluation on the long-form multi-shot video editing benchmark, MoviEdit-Bench, for edited-frame-guided propagation (IV2V).} (X-shot: cross-shot)}
    \label{tab:movie_edit_bench_genprop}
    \resizebox{\textwidth}{!}{%
    \begin{NiceTabular}{@{}>{\centering\arraybackslash}p{1.0cm}l*{16}{c}}
        \CodeBefore
        \columncolor{overallblue!22}{18}
        \Body
        \toprule
        \multirow{2}{*}{\makecell{\textbf{Edited}\\\textbf{Frame}}}
        & \multirow{2}{*}{\textbf{Model}}
        & \makecell{\textbf{Non-subject}\\\textbf{Preservation}}
        & \multicolumn{2}{c}{\makecell{\textbf{Edit}\\\textbf{Localization}}}
        & \multicolumn{3}{c}{\makecell{\textbf{Subject}\\\textbf{Preservation}}}
        & \makecell{\textbf{X-shot Edit}\\\textbf{Consistency}}
        & \multicolumn{3}{c}{\makecell{\textbf{Edit Prompt}\\\textbf{Alignment}}}
        & \makecell{\textbf{Edited-frame}\\\textbf{Fidelity}}
        & \multicolumn{4}{c}{\makecell{\textbf{Video}\\\textbf{Quality}}}
        & \makecell{\textbf{Overall Edit}\\\textbf{Success}} \\

        \cmidrule(lr){3-3}
        \cmidrule(lr){4-5}
        \cmidrule(lr){6-8}
        \cmidrule(lr){9-9}
        \cmidrule(lr){10-12}
        \cmidrule(lr){13-13}
        \cmidrule(lr){14-17}
        \cmidrule(lr){18-18}

        & & LPIPS$\downarrow$
        & Precision$\uparrow$ & Recall$\uparrow$
        & MFS$\uparrow$ & LPIPS$\downarrow$ & VLM$\uparrow$
        & VLM$\uparrow$
        & CLIP$\uparrow$ & ViCLIP$\uparrow$ & VLM$\uparrow$
        & VLM$\uparrow$
        & DINO$\uparrow$ & NIQE$\downarrow$ & PickS.$\uparrow$ & VLM$\uparrow$
        & VLM$\uparrow$ \\

        \midrule

        \multirow{4}{*}{\rotatebox[origin=c]{90}{First}}
        & HappyHorse-1.0
        & 0.244 & 0.751 & 0.828 & 0.417 & 0.304 & 0.767
        & \underline{0.925}
        & 0.220 & \textbf{0.195} & 0.841
        & 0.680
        & 0.985 & 5.08 & 19.6 & 0.707
        & 0.522 \\

        & GenProp
        & 0.0788 & \textbf{0.976} & 0.671 & 0.698 & \underline{0.104} & \underline{0.993}
        & 0.585
        & \textbf{0.232} & 0.186 & 0.761
        & 0.561
        & 0.989 & \textbf{3.96} & 19.7 & 0.789
        & 0.581 \\

        & Aleph 2.0
        & \textbf{0.0412} & 0.962 & \textbf{0.974} & \textbf{0.702} & \textbf{0.0895} & 0.985
        & 0.906
        & 0.231 & \underline{0.192} & \textbf{0.971}
        & \underline{0.819}
        & \underline{0.989} & 4.90 & \textbf{19.8} & \textbf{0.801}
        & \textbf{0.806} \\

        & \textbf{MoviEdit (Ours)}
        & \underline{0.0586} & \underline{0.965} & \underline{0.938} & \underline{0.699} & 0.107 & \textbf{1.00}
        & \textbf{0.981}
        & \underline{0.231} & 0.191 & \underline{0.938}
        & \textbf{0.828}
        & \textbf{0.989} & \underline{4.67} & \underline{19.8} & \underline{0.797}
        & \underline{0.803} \\

        \midrule

        \multirow{3}{*}{\rotatebox[origin=c]{90}{Arbitrary}}
        & HappyHorse-1.0
        & 0.248 & 0.742 & 0.793 & 0.402 & 0.307 & 0.760
        & \underline{0.913}
        & 0.219 & \textbf{0.194} & 0.813
        & 0.636
        & 0.984 & 5.11 & 19.6 & 0.709
        & 0.486 \\

        & Aleph 2.0
        & \textbf{0.0422} & \underline{0.955} & \textbf{0.952} & \textbf{0.705} & \textbf{0.0838} & \underline{0.981}
        & 0.865
        & \underline{0.229} & \underline{0.193} & \textbf{0.957}
        & \underline{0.801}
        & \underline{0.989} & \underline{4.94} & \textbf{19.8} & \underline{0.807}
        & \underline{0.781} \\

        & \textbf{MoviEdit (Ours)}
        & \underline{0.0598} & \textbf{0.963} & \underline{0.920} & \underline{0.699} & \underline{0.0996} & \textbf{0.995}
        & \textbf{0.933}
        & \textbf{0.229} & 0.192 & \underline{0.925}
        & \textbf{0.805}
        & \textbf{0.989} & \textbf{4.78} & \underline{19.7} & \textbf{0.807}
        & \textbf{0.786} \\
        \bottomrule
    \end{NiceTabular}%
    }

\end{table*}

\section{Experiments}
\subsection{Experimental Setup}

\begin{figure}
\centering
\includegraphics[width=\textwidth]{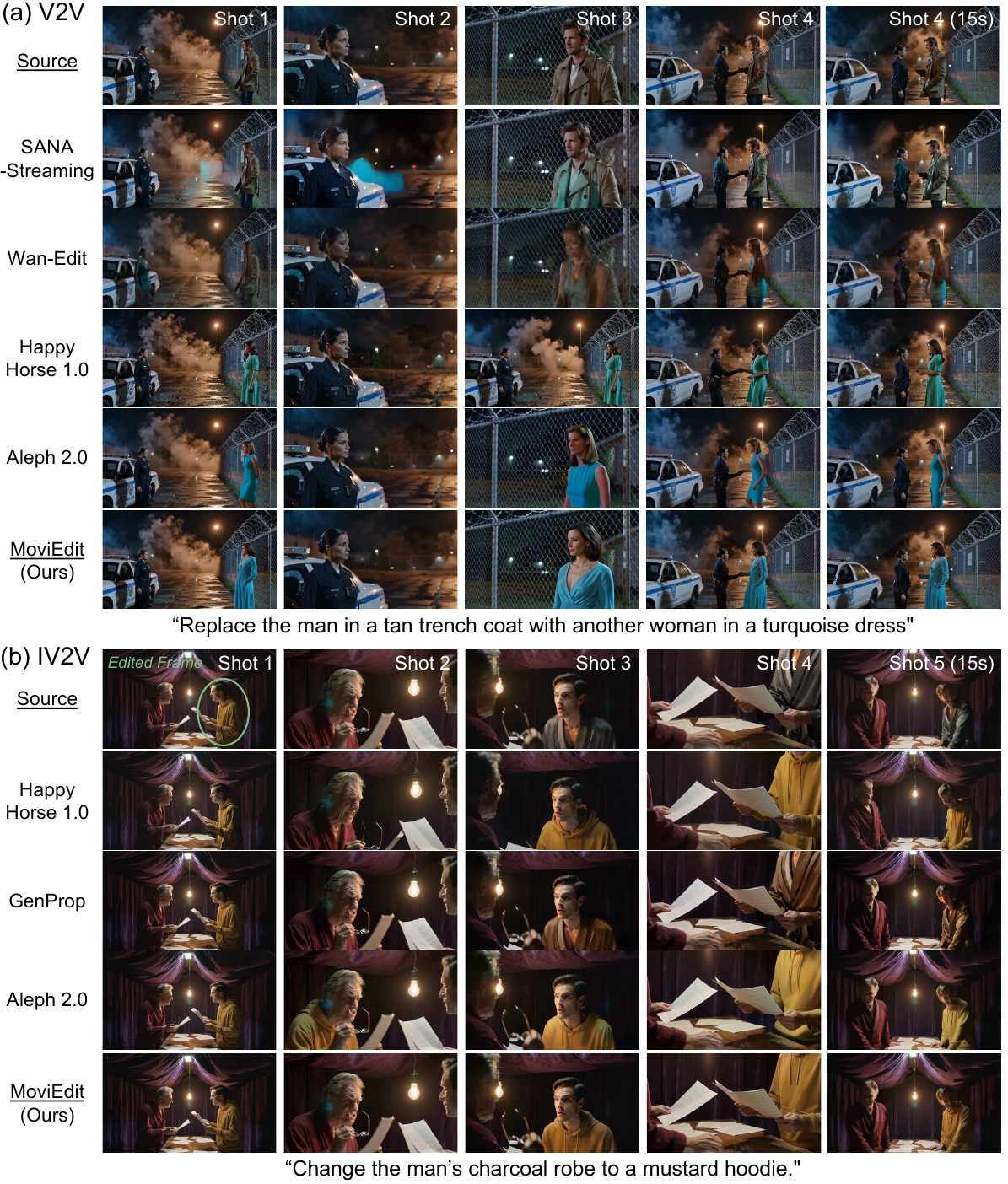}
\caption[Qualitative comparison on MoviEdit-Bench with representative baselines]{%
  \textbf{Qualitative comparison on MoviEdit-Bench with representative baselines.} 
}
\label{fig:qualitative}
\end{figure}

\begin{figure}[t]
    \centering

    \begin{minipage}{0.665\textwidth}
        \centering
        \includegraphics[width=\textwidth]{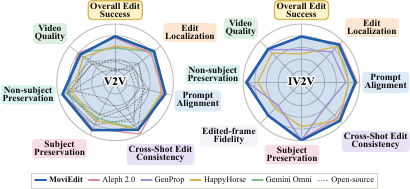}
        \caption{\textbf{MoviEdit-Bench radar chart showing that MoviEdit achieves performance on par with proprietary systems.}}
        \label{fig:radar}
    \end{minipage}
    \hfill
    \begin{minipage}{0.325\textwidth}
        \centering
        \includegraphics[width=\textwidth]{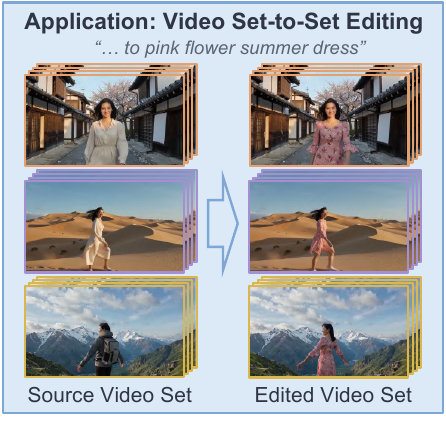}
        \caption{\textbf{MoviEdit naturally extends to video set-to-set editing setup.}}
        \label{fig:set2set}
    \end{minipage}
\end{figure}

\begin{figure}[b]
    \centering
    \vspace{1em}
    \begin{minipage}[t]{0.48\linewidth}
        \vspace{-13.5em}
        \centering

        \captionof{table}{\textbf{Training dataset statistics at 540p, 10-15s, and 4-6 shots per video.}}
        \label{tab:dataset}
        \small
        \resizebox{\linewidth}{!}{
        \begin{tabular}{lccc}
        \toprule
        \textbf{Task} & \textbf{Source} & \textbf{\# raw pairs} & \textbf{\# filtered pairs} \\
        \midrule
        Localization       & Multi-cam   & 13,801 & 10,500 \\
        Pixel-aligned      & Multi-cam & 38,979   & 21,008 \\
                           & Cinematic & 138,633 & 50,405 \\
        Semi-pixel-aligned & Generative & 13,815 & 4,706 \\
        \midrule
        \textbf{Total} & & \textbf{205,228} & \textbf{86,619} \\
        \bottomrule
        \end{tabular}
        }

        \vspace{10pt}

        \captionof{table}{
        \textbf{Inference efficiency ablation at 540p and 361 frames
        on 8 H100 GPUs (seconds).}
        }
        \label{tab:efficiency_ablation}
        \small

        \resizebox{\linewidth}{!}{
        \setlength{\tabcolsep}{1pt}
        \begin{tabular}{lccc}
        \toprule
        \textbf{Variant} &
        \textbf{Draft Editor} &
        \textbf{Render Editor} &
        \textbf{Total} \\
        \midrule
        Full temporal attn.
            & 25.1 & 339.0 & 364.1 ($\times$1.0) \\

        + Sliding-window attn.
            & - & 233.8 & 258.9 ($\times$1.4) \\

        + KV cache
            & - & 141.7 & 166.8 ($\times$2.2) \\

        + Distillation
            & \textbf{3.6}
            & \textbf{8.2}
            & \textbf{11.8} \textbf{($\times$30.9)} \\
        \bottomrule
        \end{tabular}
        }
    \end{minipage}
    \hfill
    \begin{minipage}[t]{0.50\linewidth}
        \centering

        \includegraphics[width=\linewidth]{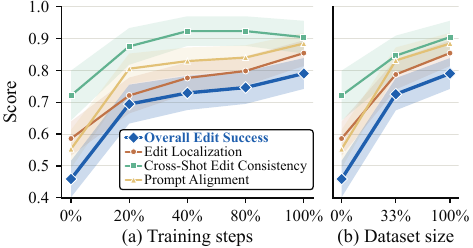}

        \caption{
        \textbf{Scaling study on (a) training steps and (b) dataset size.}
        X-axes show fractions of the full schedule (17K steps) /
        full dataset (87K pairs) used.
        Shading denotes 95\% confidence intervals.
        }
        \label{fig:scaling}

    \end{minipage}
\end{figure}

\textbf{Implementation Details.}
Using the three data-generation pipelines described above, we construct 87K paired multi-shot editing videos (540p, 15s, 24 fps) from 61,251 raw clips across three data sources (Tab.~\ref{tab:dataset}). 
In addition to this multi-shot dataset, we incorporate a proprietary collection of short, single-shot video editing pairs (540p, 5s, 24fps) during training.
Our model uses a DiT-based mixture-of-experts (MoE) backbone with a frozen multimodal text encoder and video VAE.
The input sequence concatenates text, source-video, noisy target-video, and task-specific conditioning tokens, including edited-frame and draft-edit conditions.
We initialize from a pretrained short-form, single-shot video editing checkpoint \textbf{\textit{with no prior multi-shot generation or editing capability}}.
We fully fine-tune all DiT parameters using Fully Sharded Data Parallel (FSDP) across 64 NVIDIA H200 GPUs.
Specifically, training proceeds in two stages.
We first train the video-to-video (V2V) model for 17K steps.
We then specialize the model for IV2V with an additional 8K training steps under the IV2V task formulation.
For this stage, edited reference frames are sampled from the ground-truth target video and restricted to shots in which the edited subject is present.

\textbf{Evaluations.}
We evaluate MoviEdit and competing methods on MoviEdit-Bench under both text-guided (V2V) and edited-frame-guided (IV2V) settings. 
V2V models receive only the source video and edit instruction, while IV2V models additionally receive an edited reference frame; neither setting provides object masks.
MoviEdit-Bench includes both first-frame and arbitrary-frame IV2V scenarios. 
For fair comparison, we evaluate all IV2V-capable models on the first-frame subset, and methods supporting arbitrary-frame conditioning on the full benchmark.
As no existing, fully accessible public benchmark specifically targets long-form multi-shot editing, we additionally evaluate V2V performance on the established short-form benchmarks FiVE-Bench~\citep{li2025five} and VPBench~\citep{bian2025videopainter}.
For VPBench, which is originally mask-conditioned, we evaluate MoviEdit in a mask-free setting using only the source video and edit instruction. 
We compare against various public and proprietary systems, including Lucy Edit Dev~\citep{decart2025lucyedit}, Memory-V2V~\citep{lee2026memory}, SANA-Streaming~\citep{zhao2026sana}, Wan-Edit~\citep{li2025five}, SDEdit with HoloCine~\citep{meng2022sdedit,meng2026holocine}, HappyHorse-1.0~\citep{alibaba2026happyhorse}, Aleph~2.0~\citep{runway2026aleph2}, Gemini Omni Flash~\citep{google2026geminiomni}, and GenProp~\citep{liu2025generative}.

\subsection{Results}

\textbf{Comparisons.} We compare MoviEdit with publicly available and proprietary systems on 15-second multi-shot editing using MoviEdit-Bench. For V2V, MoviEdit achieves the strongest overall performance among the evaluated systems, surpassing Runway Aleph~2.0 in overall edit success and leading on the majority of evaluation metrics (Tab.~\ref{tab:movie_edit_bench_v2v}). For IV2V, MoviEdit achieves the highest overall success on the full benchmark and performs on par with Aleph~2.0 across the individual evaluation dimensions, with complementary strengths across different metrics (Tab.~\ref{tab:movie_edit_bench_genprop}). The radar plot in Fig.~\ref{fig:radar} further highlights the balanced performance of MoviEdit across the major benchmark criteria. We further evaluate MoviEdit on the established FiVE-Bench and VPBench as external validation (Tab.~\ref{tab:external_bench_v2v}), obtaining results consistent with those observed on MoviEdit-Bench.

Fig.~\ref{fig:qualitative} shows qualitative V2V and IV2V comparisons. Public baselines exhibit pronounced failure modes, including artifacts propagated from previous shots, incorrect cross-shot edit localization, edit failure, and poor source consistency. HappyHorse and Gemini Omni generally perform the requested edits well but systematically deviate from the source, failing to preserve shot transitions and timing and hallucinating motion absent from the source. Aleph~2.0 provides the strongest baseline results, while MoviEdit shows better source preservation, edit localization, and prompt alignment. These trends are consistent with quantitative results on MoviEdit-Bench.

\textbf{Scaling and Efficiency Analysis.}
We study MoviEdit’s scaling behavior and inference efficiency.
For scaling, we separately vary training duration and dataset size: 0\%, 20\%, 40\%, 80\%, and 100\% of the 17K-step schedule, and 0\%, 33\%, and 100\% of the 87K-pair dataset, respectively.
Increasing either dataset size or training duration consistently improves performance across metrics (Fig.~\ref{fig:scaling}), suggesting further gains from scaling. 
We next ablate inference efficiency at 540p and 361 frames on 8 H100 GPUs (Tab.~\ref{tab:efficiency_ablation}). 
Starting from full temporal attention (364.1\,s), sliding-window attention reduces denoising latency to 258.9\,s, and KV caching further reduces it to 166.8\,s. 
Combined with few-step distillation, the full system runs in 11.8\,s, yielding a $30.9{\times}$ speedup over the full-attention baseline. 
Together, these optimizations reduce the denoising latency to 11.8\,s, below the 15\,s input-video duration, substantially improving the practicality of long-form multi-shot editing.

\begin{figure*}[t]
    \centering
    \includegraphics[width=\textwidth]{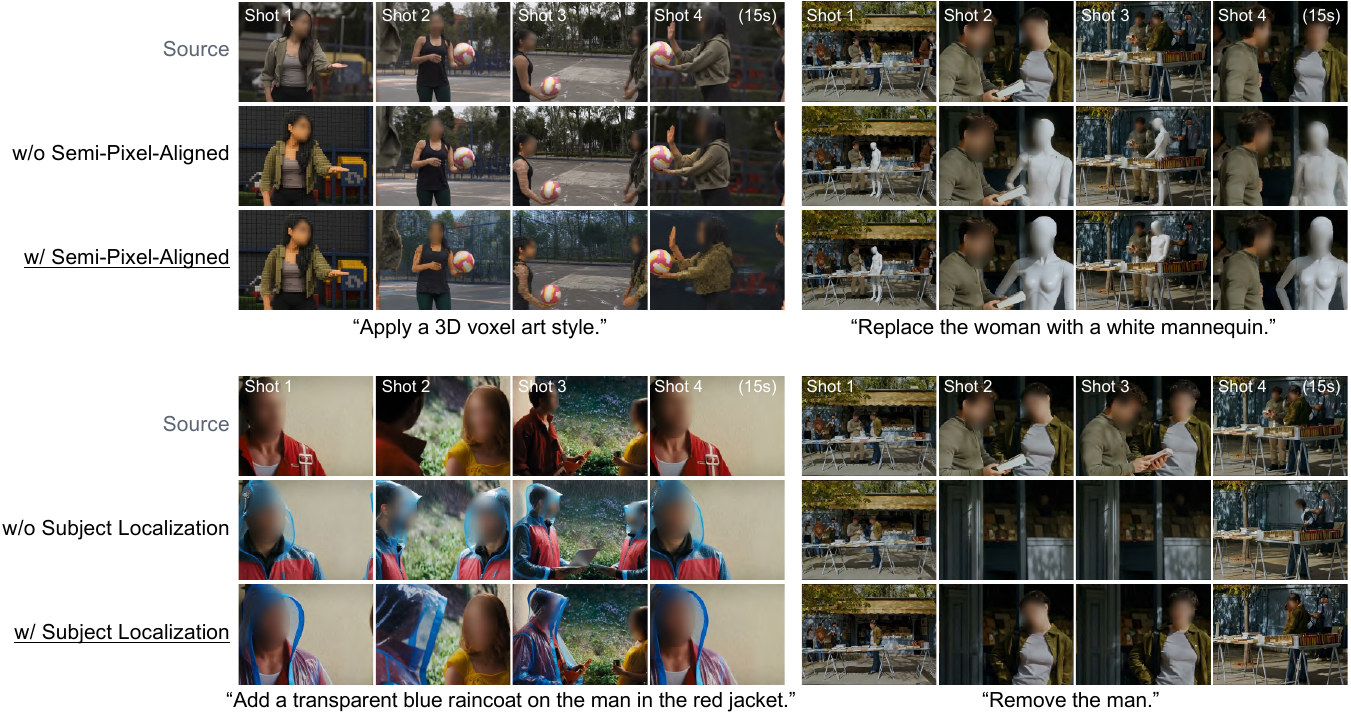}
    \caption{
    \textbf{Ablation of task-specific data recipes.}
    \textbf{Top:} Adding semi-pixel-aligned training data improves generalization to edits involving relatively larger changes in structure or shape.
    \textbf{Bottom:} Adding subject localization training data improves the model's ability to identify the intended edit target, leading to improved edit localization.
    }
    \label{fig:data_ablation}
\end{figure*}

\textbf{Data Recipe Ablations.}
To examine the effect of our task-specific data recipes, we conduct small-scale ablation training by selectively removing different supervision pipelines.
Fig.~\ref{fig:data_ablation} shows qualitative comparisons for subject-localization and semi-pixel-aligned training data.
Semi-pixel-aligned data improves generalization to edits that require relatively larger changes in structure or shape.
As shown in the top part of Fig.~\ref{fig:data_ablation}, the model trained with semi-pixel-aligned data better handles such edits.
This complements our pixel-aligned data pipeline, which has difficulty constructing source--target editing pairs involving large structural or shape changes.
Adding subject localization data improves the model's ability to identify the intended edit target.
As shown in the bottom part of Fig.~\ref{fig:data_ablation}, the model trained with subject localization supervision exhibits improved edit localization compared with the variant trained without it. This observation is consistent with prior findings in video editing~\citep{liu2025generative}, where task-specific tracking supervision has been shown to benefit editing performance.

\textbf{Applications.}
We explore applications beyond standard V2V and IV2V multi-shot editing, including settings not seen during training. 
First, MoviEdit supports \textit{multi-reference-frame editing} (Fig.~\ref{fig:title}(b)); although trained with a single edited reference frame, it can condition on multiple edited references at inference time, combine edits to different subjects, and propagate them consistently.
Second, we extend image set-to-set editing~\citep{feingold2026match} to videos with \textit{video set-to-set editing} (Fig.~\ref{fig:set2set}).
We concatenate multiple source videos into a single multi-shot sequence and edit them jointly, yielding consistent transformations despite variations in appearance and background.
Finally, we evaluate \textit{beyond-training-horizon editing} (Fig.~\ref{fig:title}(c)). 
Although trained on videos up to 15 seconds, MoviEdit generalizes to 25-second inputs with up to 17 shots, as well as long-take single-shot videos, extending both temporal duration and shot count beyond the training regime.

\begin{figure}[!t]
    \centering
    \includegraphics[width=0.99\textwidth]{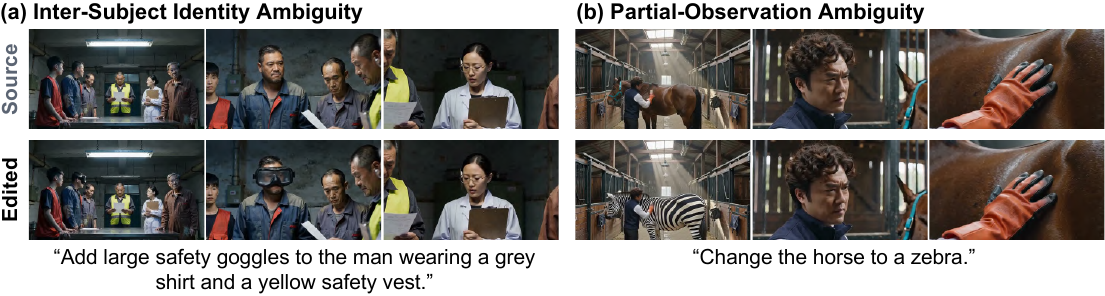}
    \caption{
    \textbf{Failure cases of MoviEdit.}
    (a) \textbf{Inter-subject identity ambiguity.}
    When multiple visually similar subjects are present, the model may mistakenly apply the edit to an unintended subject rather than the specified target.
    (b) \textbf{Partial-observation ambiguity.}
    When the target is only partially visible, the model may struggle to accurately localize and edit the intended subject.
    }
    \label{fig:supp-fail}
\end{figure}

\textbf{Limitations and Failure Cases.}
Fig.~\ref{fig:supp-fail} highlights two challenging target-localization failure modes.
In Fig.~\ref{fig:supp-fail}(a), multiple visually similar subjects create identity ambiguity, which can cause the edit to be applied to the wrong target.
In Fig.~\ref{fig:supp-fail}(b), partial target visibility (\textit{e.g.}, only a body part or object fragment is visible) can make cross-shot association difficult, leading to incomplete or incorrect edits.

\section{Conclusion}

We presented MoviEdit, a unified framework for long-form multi-shot video editing that jointly addresses scalable supervision, efficient modeling, and dedicated evaluation.
MoviEdit combines scalable pipelines for generating paired multi-shot editing data, cascaded bidirectional--autoregressive modeling, and MoviEdit-Bench for evaluating cross-shot localization, consistency, preservation, and edit alignment. 
Across text- and image-guided editing, MoviEdit substantially outperforms existing open-source models and matches strong proprietary systems, while generalizing to set-to-set editing and videos beyond the training horizon. 
Regarding limitations, MoviEdit can struggle with target localization under strong identity ambiguity, such as when multiple visually similar subjects are present or the target is only partially visible.
These limitations motivate several directions for future work.
Agentic editing workflows could further improve edit fidelity and computational efficiency, for example by selectively processing only the shots or subjects that require editing.
On the data side, controllable rendering engines \citep{engine2018unreal} offer another scalable source of supervision.
Finally, MoviEdit focuses on correspondence-preserving edits that retain the source motion and narrative structure; extending multi-shot long video editing to free-form transformations of motion and narrative structure remains an important future direction.

\bibliography{iclr2027_conference}
\bibliographystyle{iclr2027_conference}

\clearpage

\appendix

\section*{Supplementary Material}

This supplementary material provides additional details and results
to complement the main paper. We first provide additional
details on our training dataset construction (Sec.~\ref{sec:supp-data-source}) and MoviEdit-Bench (Sec.~\ref{sec:supp-bench}),
including its design and evaluation metrics. 
Finally, we present additional experimental results on external benchmark comparisons (Sec.~\ref{sec:supp-external-bench}) and beyond-training-horizon editing (Sec.~\ref{sec:supp-very-long}).

\section{Training Dataset Construction} \label{sec:supp-data-source}

\begin{figure}[!b]
    \centering
\includegraphics[width=1.0\textwidth]{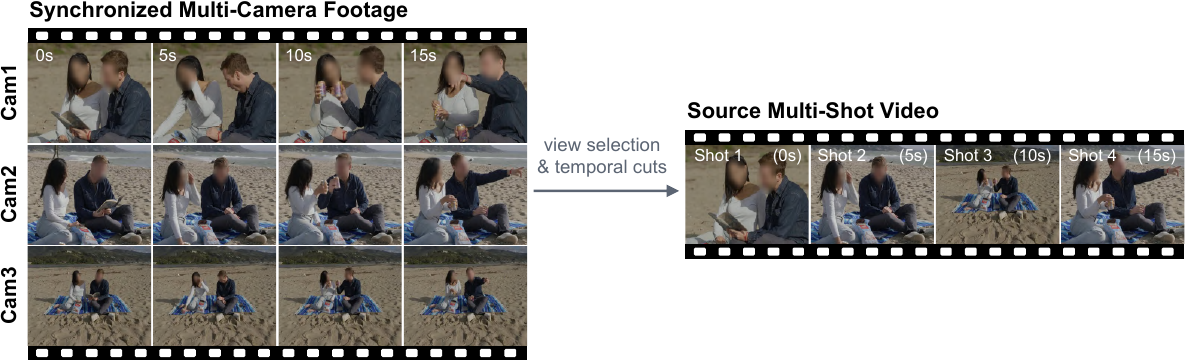}
\caption{
\textbf{Examples of synchronized multi-camera footage used as raw data for MoviEdit.} Each scene is captured simultaneously from 2--4 cameras, showing the same real-world scene from different viewpoints while actors perform natural interactions. \textit{Note: Faces of real actors in the displayed examples are blurred to protect their identity.}
}
\label{fig:supp-data-source}
\end{figure}
 
\textbf{Raw Data Sources.}
We collect raw video clips from three sources: synchronized multi-camera footage (1,147 clips), cinematic footage (32,474 clips), and generative videos (27,630 clips). In particular, the synchronized multi-camera footage captures the same real-world scene simultaneously from 2--4 cameras, with actors performing natural interactions. This setup is conceptually similar to the synthetic SynCamMaster setup \citep{bai2025syncammaster}, but uses real-world footage. Representative examples are shown in Fig.~\ref{fig:supp-data-source}.

\textbf{Editing Pair Generation.}
Using these raw data sources, we construct the MoviEdit training dataset through three data-generation pipelines: localization, pixel-aligned, and semi-pixel-aligned. The resulting dataset contains 86,619 paired multi-shot videos at 540p resolution, approximately 15 seconds in duration, and 24 fps. The detailed breakdown by pipeline, data source, and editing task is provided in Table~\ref{tab:supp-data}.

\begin{table*}[t]
\centering
\small
\setlength{\tabcolsep}{6pt}
\renewcommand{\arraystretch}{1.1}

\resizebox{0.6\textwidth}{!}{%
\begin{tabular}{lllr}
\toprule
\textbf{Pipeline} &
\textbf{Data source} &
\textbf{Editing task} &
\textbf{\# Pairs} \\
\midrule

\multirow{3}{*}{Localization}
& \multirow{3}{*}{Multi-camera}
& Highlighting  & 3,500 \\
&
& Outlining     & 3,500 \\
&
& Color filling & 3,500 \\
\cmidrule(lr){3-4}
&
& \textit{Subtotal} & 10,500 \\

\midrule

\multirow{12}{*}{Pixel-aligned}
& \multirow{6}{*}{Multi-camera}
& Add                & 2,162 \\
&
& Remove             & 3,209 \\
&
& Appearance change  & 4,530 \\
&
& Swap               & 2,926 \\
&
& Stylization        & 3,841 \\
&
& Environment change & 4,340 \\
\cmidrule(lr){3-4}
&
& \textit{Subtotal}  & 21,008 \\
\cmidrule(lr){2-4}

& \multirow{6}{*}{Cinematic}
& Add                & 6,787 \\
&
& Remove             & 11,921 \\
&
& Appearance change  & 14,206 \\
&
& Swap               & 8,742 \\
&
& Stylization        & 1,606 \\
&
& Environment change & 7,143 \\
\cmidrule(lr){3-4}
&
& \textit{Subtotal}  & 50,405 \\

\midrule

\multirow{3}{*}{Semi-pixel-aligned}
& \multirow{3}{*}{Generative}
& Environment change          & 1,364 \\
&
& Medium change        & 1,368 \\
&
& Character appearance change & 1,974 \\
\cmidrule(lr){3-4}
&
& \textit{Subtotal}           & 4,706 \\

\midrule
\multicolumn{3}{r}{\textbf{Total}} &
\textbf{86,619} \\
\bottomrule
\end{tabular}
}
\caption{
\textbf{Detailed composition of the MoviEdit training dataset by data-generation pipeline, data source, and editing task.}
}
\label{tab:supp-data}
\end{table*}

\section{MoviEdit-Bench} \label{sec:supp-bench}

\subsection{Source Multi-Shot Videos} MoviEdit-Bench comprises 64 source videos, primarily covering cinematic, story-driven multi-shot scenes with coherent space, time, and subject identity across shots. The scenes span diverse settings, including single-subject activities, interpersonal interactions, larger ensembles, and human--animal interactions, with deliberate variation in pose, viewpoint, visibility, and camera framing to support cross-shot editing evaluation.

The source videos are organized into eight high-level categories (Tab.~\ref{tab:supp-bench-src-cat}). Categories C1--C6 progressively increase the difficulty of grounding an edit instruction to the correct subject across shots, ranging from a single persistent subject to crowded multi-subject and human--animal scenes. C3 is further divided into same-gender and different-gender two-person settings to examine different levels of subject ambiguity. C7 evaluates robustness across non-photorealistic visual media, while C8 targets scene-level global edits without entity localization.

Each scenario is authored by an LLM as a structured specification containing a global scene description, per-shot descriptions, and associated metadata. We generate each source video from scratch using a multi-shot text-to-video model, avoiding reliance on copyrighted source footage. Each video contains four to six shots and 361 frames at 24 fps, corresponding to approximately 15 seconds. The prompts, metadata, and generated videos are then manually reviewed for scene coherence, subject identity, shot correspondence, and suitability for editing evaluation.

\begin{table*}[t]
\centering
\small
\setlength{\tabcolsep}{4pt}
\renewcommand{\arraystretch}{1.15}
\resizebox{\textwidth}{!}{%
\begin{tabular}{
    p{0.09\textwidth}
    p{0.18\textwidth}
    c
    p{0.36\textwidth}
    p{0.27\textwidth}
}
\toprule
\textbf{Category} &
\textbf{Composition} &
\textbf{\# Videos} &
\textbf{Source-video design} &
\textbf{Primary evaluation focus} \\
\midrule

C1 &
Single human &
6 &
One persistent subject observed across varying actions, poses, framings, and viewpoints. &
Cross-shot localization under pose and viewpoint changes. \\

C2 &
Two isolated humans &
6 &
Two humans belong to the same scene, while each shot contains only one of them. &
Target selection across shots and correct exclusion when the target is absent. \\

C3A &
Two co-visible humans, same gender &
6 &
Same-gender subjects are distinguished by roles, clothing, hairstyles, accessories, positions, or object ownership. &
Subject disambiguation between visually and semantically competing humans. \\

C3B &
Two co-visible humans, different gender &
6 &
Different-gender subjects co-occur within shots and are distinguished by roles and visual attributes. &
Cross-shot subject binding with multiple co-visible candidates. \\

C4 &
Three humans &
6 &
Three text-addressable characters share a continuous scene with distinct visual or semantic identifiers. &
Localization under increased multi-person competition. \\

C5 &
Four-or-more-person ensemble &
6 &
Larger ensembles contain distinguishable roles, appearance, props, and spatial positions. &
Target localization in crowded multi-person scenes. \\

C6 &
Human and animal &
6 &
A human and an animal coexist across a continuous multi-shot scene. &
Cross-species target grounding and identity persistence. \\

C7 &
Visual-medium robustness &
6 &
Scenes span non-photorealistic media such as stylized 3D, stop-motion, watercolor, claymation, anime, and paper-cut styles. &
Robustness of localization and editing across visual media. \\

C8 &
Scene-level global edits &
16 &
Scenes are designed for global transformations such as stylization, lighting, weather, and time-of-day changes. &
Global editing without subject-localization constraints. \\

\midrule
\multicolumn{2}{r}{\textbf{Total}} &
\textbf{64} &
&
\\
\bottomrule
\end{tabular}
}
\caption{
\textbf{Overview of the MoviEdit-Bench source-video categories.} C1--C6 progressively vary cross-shot subject-grounding difficulty, C7 evaluates robustness across visual media, and C8 targets scene-level global editing.
}
\label{tab:supp-bench-src-cat}
\end{table*}

\subsection{Editing Scenarios}

We construct two editing scenarios for each source video, yielding 128 scenarios in total. The 48 entity-centric videos in C1--C7 provide local edits spanning \textit{add}, \textit{remove}, \textit{appearance change}, and \textit{swap}, with a 3:1 ratio between subject targets (characters or animals) and object targets. The 16 videos in C8 provide scene-level global edits covering \textit{environment change} and \textit{stylization}. Overall, local and global edits account for 75\% and 25\% of the benchmark, respectively; the detailed task distribution is given in Tab.~\ref{tab:supp-bench-task-dist}.

Editing scenarios and instructions are authored by an LLM from the source-video descriptions and structured entity metadata, following the predefined task distribution. All scenarios and edit instructions are subsequently manually reviewed for target correctness, instruction clarity, and suitability for evaluation.

For edited-frame-guided propagation, MoviEdit-Bench supports arbitrary edited-frame conditioning. We divide the 128 scenarios evenly into 64 first-frame and 64 non-first-frame reference settings. For each scenario, a suitable reference frame is manually selected, ensuring visibility of the edited target when applicable, and edited with an off-the-shelf text-guided image editor using the corresponding edit instruction.

\begin{table}[t]
\centering
\small
\setlength{\tabcolsep}{5pt}
\renewcommand{\arraystretch}{1.1}
\resizebox{0.45\textwidth}{!}{%
\begin{tabular}{lccc}
\toprule
\textbf{Edit task} & \textbf{Scope} & \textbf{\# Scenarios} & \textbf{Ratio} \\
\midrule
Add                & Local  & 24 & 18.75\% \\
Remove             & Local  & 24 & 18.75\% \\
Appearance change  & Local  & 24 & 18.75\% \\
Swap               & Local  & 24 & 18.75\% \\
Environment change & Global & 24 & 18.75\% \\
Stylization        & Global & 8  & 6.25\% \\
\midrule
\textbf{Total}     &        & \textbf{128} & \textbf{100\%} \\
\bottomrule
\end{tabular}
}
\caption{
\textbf{Distribution of the 128 MoviEdit-Bench editing scenarios across local and global edit categories.}
}
\label{tab:supp-bench-task-dist}
\end{table}

\subsection{Evaluation Metadata}

Beyond the inputs required for benchmark evaluation, including the source video, edit instruction, and edited reference frame for the IV2V task, MoviEdit-Bench provides additional metadata for metric computation. For local edits, we provide per-frame masks of the editing target, obtained using video segmentation models~\citep{cheng2024putting, sari2025flowcut, carion2026sam} followed by manual verification and correction. The masks are temporally aligned with the source video and paired with the corresponding edit instruction. We also provide shot-boundary metadata for each source video, represented as the start and end frame indices of each detected shot using OmniShotCut~\citep{wang2026omnishotcut} for shot-level metric computation.

Each editing scenario is further accompanied by a target-video prompt. Following the same format as the source-video generation prompt, it contains a global scene description together with per-shot descriptions. The target prompt is authored as a self-contained description of the fully edited target video, rather than as a relative modification of the source prompt, and is manually reviewed for consistency with the intended edit. These target prompts are also used to evaluate baselines formulated with target-video conditioning rather than standard edit instructions, such as Wan-Edit~\citep{li2025five} and SDEdit~\citep{meng2022sdedit}.

\subsection{Evaluation Metrics}
\label{sec:supp-metrics}

We evaluate each edited video along seven complementary dimensions: \textit{non-subject preservation}, \textit{edit localization}, \textit{subject preservation}, \textit{cross-shot edit consistency}, \textit{edit-prompt alignment}, \textit{video quality}, and \textit{edited-frame fidelity}. We additionally report an \textit{overall success} score that jointly considers multiple editing criteria. Each metric produces one score per editing scenario, and model-level results are averaged over applicable scenarios. Metric applicability varies by edit type, as summarized in Tab.~\ref{tab:supp-bench-yes-or-no}.

\textbf{Evaluation protocol.}
Methods may differ in output resolution and duration. We resize the source video, edited video, and target masks to a common $960\times544$ resolution and temporally align them to the 361-frame source video. When a method produces fewer frames, we evaluate only the overlapping range. The shot boundaries provided with MoviEdit-Bench define the units for shot-level evaluation, and temporal computations are performed independently within each shot. For mask-based metrics, we dilate the target masks by 9 pixels to exclude ambiguous boundary regions and omit frames or shots in which the target is not visible.

\textbf{VLM judge.}
We use a VLM~\citep{team2026gemma, yang2025qwen3} to evaluate semantic properties that are difficult to capture with pixel- or embedding-level similarity, such as whether an edit is applied to the correct subject. Unless otherwise specified, VLM-based metrics share a common frame-pair protocol. We uniformly sample four frames from each shot, including the first and last frames. For each sample, the judge receives the source frame, edited frame, edit instruction, and target description, and independently determines whether: (i) the target is visible in the source frame, (ii) the requested edit is present, (iii) the edit is applied to the intended target, (iv) unrelated content is preserved, (v) target attributes unrelated to the instruction are preserved, and (vi) the edit completely covers the target. Frames are sampled from all shots, including those in which the target is absent, enabling the judge to detect edits that incorrectly leak to other subjects or shots. For global edits, the corresponding checks assess whether the requested scene-level change is present, covers the entire scene, and preserves the underlying scene content. All VLM metrics below reuse these judge outputs unless otherwise noted. Each binary VLM verdict is mapped to 1 for \texttt{true} and 0 for \texttt{false}. Frame-level VLM metrics are averaged over all applicable sampled frames, while scenario-level judgments, such as cross-shot consistency, directly yield a binary scenario score.

\textbf{Non-subject preservation.}
This dimension measures preservation outside the intended edit region. For local edits, we compute LPIPS~\citep{zhang2018unreasonable} between corresponding source and edited frames after masking out the dilated target region, and average the score over frames.

\textbf{Edit localization.}
This dimension measures whether the edit is confined to the intended target (\textit{Precision}) and sufficiently covers it (\textit{Recall}). Precision is the fraction of sampled frames passing VLM check~(iv), penalizing unintended changes to competing subjects or to shots in which the target is absent. Recall is the fraction passing check~(vi), measuring whether the intended edit is completed rather than only partially applied. For global edits, Recall instead measures whether the requested change covers the entire scene, while Precision is undefined.

\textbf{Subject preservation.}
This dimension measures preservation of target properties that are not requested to change. \textit{MFS} follows the Motion Fidelity Score~\citep{yatim2024space} used in FiVE-Bench~\citep{li2025five}, adapted to multi-shot videos. We compute it independently for each target-bearing shot and average across shots. Within each shot, a $20\times20$ grid of points inside the source target bounding box is tracked with CoTracker3~\citep{karaev2025cotracker3} in both source and edited videos, and the normalized frame-to-frame displacement directions are compared. Since MFS evaluates motion rather than appearance, it also applies to swaps and global edits. For global edits, we instead sample tracking points over the entire frame.
\textit{LPIPS}~\citep{zhang2018unreasonable} compares the target regions between source and edited frames and is used for add and appearance-change edits, where the original subject identity should remain. \textit{VLM} reports the fraction of frames passing check~(v); for global edits, it measures preservation of the underlying scene content.

\textbf{Cross-shot edit consistency.}
This dimension measures whether the requested edit remains consistent across shots. We jointly provide the VLM judge with edited frames sampled from all target-bearing shots and ask whether visible instances exhibit the same edit, such as the same added object, modified appearance, or substituted identity. The judge is instructed to assess consistency independently of edit correctness, which is evaluated by other metrics. This produces one binary score per scenario and is defined when at least two relevant shots are available. It does not apply to removal edits. For global edits, the metric evaluates whether the resulting environment or visual style remains consistent across shots.

\textbf{Edited-frame fidelity.}
For the IV2V task, this dimension measures whether the propagated edit remains faithful to the edited reference frame. We include the reference frame in each frame-pair VLM query and evaluate whether the edited subject or scene preserves its reference identity and attributes while allowing natural variations in viewpoint, pose, lighting, and motion. The score is the fraction of positive judgments and is not defined for removal edits.

\textbf{Edit prompt alignment.}
This dimension measures alignment between the edited video and the per-shot target descriptions provided in the evaluation metadata. Each shot is evaluated against its corresponding target description, including shots that should remain unchanged. For \textit{CLIP}, we uniformly sample eight frames per shot, compute cosine similarity between each frame and its target description using CLIP ViT-L/14~\citep{radford2021learning}, and average the scores. For \textit{ViCLIP}~\citep{wang2024internvid}, the same eight frames are encoded jointly as a video clip and compared with the text embedding of the target description. \textit{VLM} reports the fraction of sampled frames passing check~(ii).

\textbf{Video quality.}
This dimension evaluates the edited video independently of edit correctness. \textit{DINO} measures temporal coherence as the mean cosine similarity between DINOv2~\citep{oquab2023dinov2} features of adjacent frames within each shot. \textit{NIQE}~\citep{mittal2012making} measures no-reference image quality and is averaged over frames, with lower values indicating better quality. \textit{PickScore}~\citep{kirstain2023pick} scores each frame against the corresponding per-shot target description. For \textit{VLM}, we jointly provide the middle frame of each shot and ask the judge to rate visual quality from 1 to 5 based on naturalness and artifacts such as warping, blur, malformed regions, unnatural lighting, and geometry. We normalize the resulting score $s$ to $[0,1]$ as $(s-1)/4$.

\textbf{Overall edit success.}
Overall edit success provides a strict frame-level measure that combines the applicable VLM criteria. A sampled frame is considered successful only when the requested edit is present, applied to the intended target, and unrelated content is preserved. When applicable, it must additionally preserve non-edited target attributes, fully cover the target, and remain consistent with the edited reference frame. The final score is the fraction of successful frames. Since it reuses the frame-pair judgments above, no additional VLM queries are required.

\begin{figure}[t]
    \centering
    \hspace*{0.1\textwidth}
\includegraphics[width=0.9\textwidth]{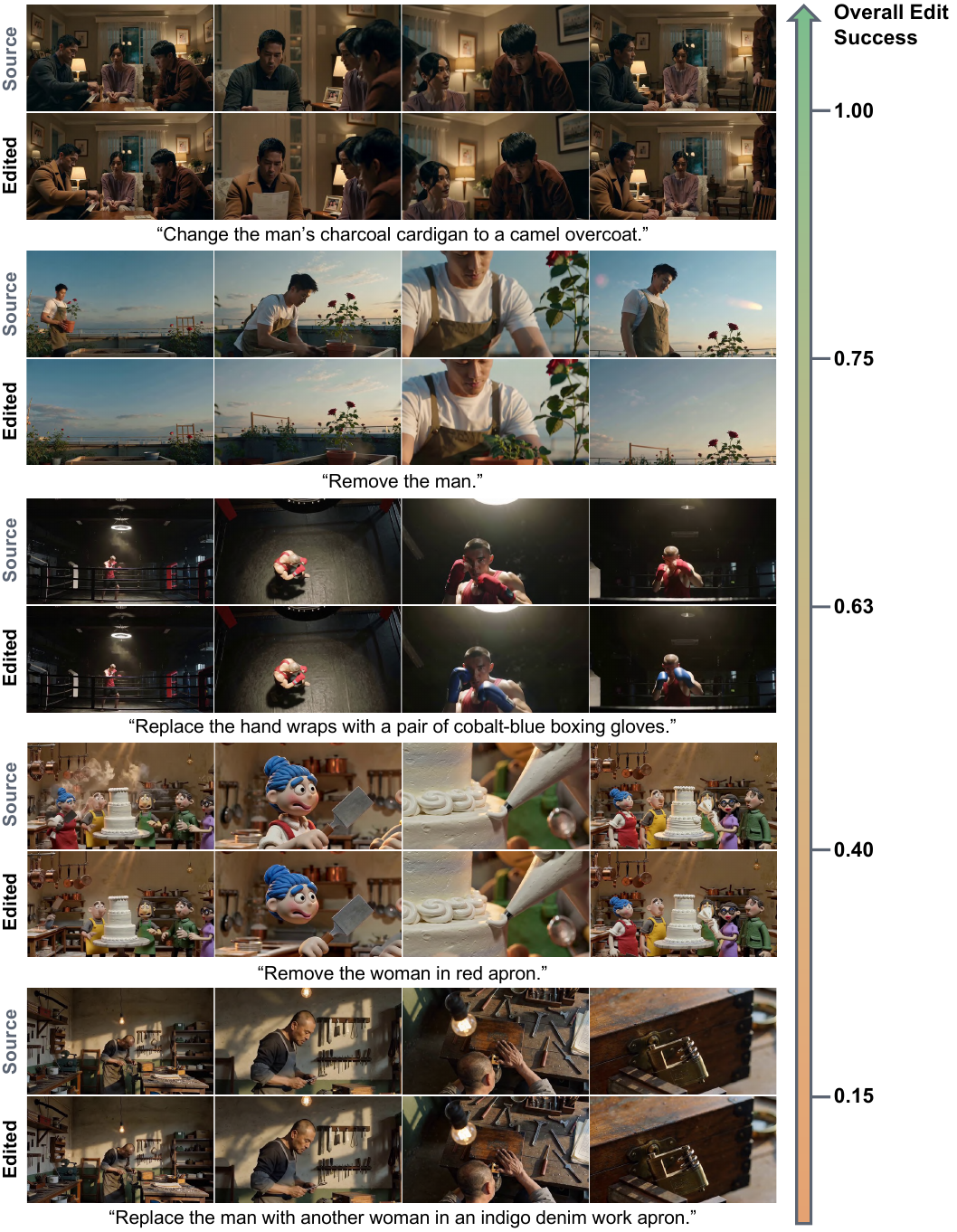}
\caption{
\textbf{Qualitative comparison across overall edit success scores.}
Representative V2V results from multiple models, ordered by overall edit success. 
Examples progress from incorrect localization or incomplete edits to successful, cross-shot-consistent editing.
}
\label{fig:supp-metric-grad}
\end{figure}

We visualize editing results from multiple models across different score ranges (Fig.~\ref{fig:supp-metric-grad}). The examples show a progression from unsuccessful edits with incorrect subject localization or incomplete edit execution to successful edits with consistent appearance across shots. 

\begin{table}[t]
\centering
\small
\setlength{\tabcolsep}{4pt}
\renewcommand{\arraystretch}{1.1}
\resizebox{0.8\textwidth}{!}{%
\begin{tabular}{llcccccc}
\toprule
& & \multicolumn{4}{c}{\textbf{Local}} & \multicolumn{2}{c}{\textbf{Global}} \\
\cmidrule(lr){3-6} \cmidrule(lr){7-8}
\textbf{Dimension} & \textbf{Metric} & Add & Remove & App. & Swap & Env. & Style \\
\midrule
Non-subject preservation & LPIPS                   & \checkmark & \checkmark & \checkmark & \checkmark & -- & -- \\
\midrule
\multirow{2}{*}{Edit localization}
                         & Precision (VLM)         & \checkmark & \checkmark & \checkmark & \checkmark & -- & -- \\
                         & Recall (VLM)            & \checkmark & \checkmark & \checkmark & \checkmark & \checkmark & \checkmark \\
\midrule
\multirow{3}{*}{Subject preservation}
                         & MFS                     & \checkmark & -- & \checkmark & \checkmark & \checkmark & \checkmark \\
                         & LPIPS                   & \checkmark & -- & \checkmark & -- & -- & -- \\
                         & VLM                     & \checkmark & -- & \checkmark & -- & \checkmark & \checkmark \\
\midrule
Cross-shot edit consistency   & VLM                     & \checkmark & -- & \checkmark & \checkmark & \checkmark & \checkmark \\
Edited-frame fidelity$^\dagger$
                         & VLM                     & \checkmark & -- & \checkmark & \checkmark & \checkmark & \checkmark \\
\midrule
Edit prompt alignment         & CLIP, ViCLIP, VLM       & \checkmark & \checkmark & \checkmark & \checkmark & \checkmark & \checkmark \\
Video quality            & DINO, NIQE, PickS., VLM & \checkmark & \checkmark & \checkmark & \checkmark & \checkmark & \checkmark \\
Overall edit success          & VLM                     & \checkmark & \checkmark & \checkmark & \checkmark & \checkmark & \checkmark \\
\bottomrule
\end{tabular}
}
\caption{
\textbf{Applicability of the evaluation metrics across edit types.} A metric is undefined (--) when its underlying premise does not apply. $^\dagger$Reported only for edited-frame-guided methods.
}
\label{tab:supp-bench-yes-or-no}
\end{table}

\begin{table*}[b]
    \centering
    \setlength{\tabcolsep}{0pt} %
    \renewcommand{\arraystretch}{1.05}
    \scriptsize
    \caption{\textbf{Quantitative evaluation on the short-form single-shot video editing benchmarks FiVE-Bench and VPBench for text-guided video editing (V2V).}}
    \label{tab:external_bench_v2v}
    \resizebox{\textwidth}{!}{%
    \begin{NiceTabular}{@{}>{\centering\arraybackslash}p{0.5cm}l*{17}{c}}
        \toprule
        & & \multicolumn{10}{c}{\textbf{FiVE-Bench}} & \multicolumn{7}{c}{\textbf{VPBench}} \\
        \cmidrule(lr){3-12} \cmidrule(lr){13-19}
        \multirow{2}{*}{} & \multirow{2}{*}{\textbf{Model}} & \textbf{Structure} & \multicolumn{4}{c}{\textbf{Background Preservation}} & \multicolumn{2}{c}{\textbf{Text Alignment}} & \textbf{IQA} & \textbf{Motion} & \textbf{FiVE-Acc} & \multicolumn{4}{c}{\textbf{Background Preservation}} & \multicolumn{2}{c}{\textbf{Text Alignment}} & \textbf{Quality} \\
        \cmidrule(lr){3-3} \cmidrule(lr){4-7} \cmidrule(lr){8-9} \cmidrule(lr){10-10} \cmidrule(lr){11-11} \cmidrule(lr){12-12} \cmidrule(lr){13-16} \cmidrule(lr){17-18} \cmidrule(lr){19-19}
        & & Dist.$\downarrow$ & PSNR$\uparrow$ & LPIPS$\downarrow$ & MSE$\downarrow$ & SSIM$\uparrow$ & CLIPS.$\uparrow$ & CLIPS.\textsubscript{edit}$\uparrow$ & NIQE$\downarrow$ & Fidelity$\uparrow$ & VLM$\uparrow$ & PSNR$\uparrow$ & SSIM$\uparrow$ & LPIPS$\downarrow$ & MSE$\downarrow$ & CLIPS.$\uparrow$ & CLIPS.\textsubscript{edit}$\uparrow$ & FVID$\downarrow$ \\
        \midrule
        \multirow{3}{*}{\rotatebox[origin=c]{90}{Open}}
        & Lucy Edit
        & 15.49 & \underline{25.66} & 99.65 & 50.52 & 82.15 & 25.41 & 20.33 & 6.30 & 87.95 & 0.3532
        & 26.53 & 0.856 & 0.091 & 0.0095 & 11.11 & \underline{13.89} & \underline{0.236} \\

        & SANA-Streaming
        & \underline{14.63} & 24.25 & 111.66 & 56.80 & 80.52 & 24.82 & 19.78 & 6.22 & 86.44 & 0.1921
        & 23.34 & 0.825 & 0.108 & 0.0100 & \underline{11.96} & 13.86 & 0.252 \\

        & Wan-Edit
        & \textbf{11.36} & 25.38 & \underline{96.88} & \underline{34.92} & 81.71 & 26.64 & 21.18 & 6.16 & 88.20 & 0.4010
        & \underline{27.09} & \underline{0.877} & \textbf{0.076} & \textbf{0.0023} & 11.00 & 13.54 & 0.238 \\

        \midrule
        \multirow{3}{*}{\rotatebox[origin=c]{90}{Private}}
        & Gemini Omni Flash
        & 31.29 & 20.45 & 132.49 & 129.43 & 67.96 & \textbf{28.02} & \underline{22.24} & \underline{6.09} & 82.27 & \textbf{0.8007}
        & 23.44 & 0.789 & 0.102 & 0.0114 & 11.46 & 13.30 & 0.270 \\

        & HappyHorse-1.0
        & 34.36 & 19.71 & 145.69 & 142.78 & 64.12 & \underline{27.85} & \textbf{22.30} & \textbf{5.88} & 83.67 & \underline{0.7864}
        & 21.19 & 0.695 & 0.126 & 0.0138 & 10.22 & 12.92 & 0.293 \\

        & Aleph 2.0
        & 23.85 & 25.00 & 104.51 & 85.82 & \underline{83.68} & 27.56 & 21.97 & 6.12 & \underline{88.69} & 0.7784
        & 26.41 & \underline{0.877} & 0.094 & 0.0121 & 10.74 & 13.36 & 0.248 \\

        \midrule
        & \textbf{MoviEdit (Ours)}
        & 16.86 & \textbf{26.59} & \textbf{88.23} & \textbf{31.73} & \textbf{84.67} & 26.95 & 21.57 & 6.32 & \textbf{89.07} & 0.6623
        & \textbf{28.30} & \textbf{0.885} & \underline{0.082} & \underline{0.0043} & \textbf{12.76} & \textbf{14.00} & \textbf{0.197} \\
        \bottomrule
    \end{NiceTabular}%
    }
\end{table*}

\section{Additional Results}
\subsection{External Benchmark Validation}
\label{sec:supp-external-bench}
To further validate MoviEdit beyond the multi-shot setting, we evaluate it on two established short-form video editing benchmarks, FiVE-Bench~\citep{li2025five} and VPBench~\citep{bian2025videopainter}. As shown in Tab.~\ref{tab:external_bench_v2v}, MoviEdit delivers performance that is comparable to or competitive with proprietary models across both benchmarks. These results complement our evaluation on MoviEdit-Bench, confirming that the proposed framework retains strong editing capabilities in conventional single-shot settings.

\subsection{Beyond-Training-Horizon Editing} \label{sec:supp-very-long}

\begin{figure}[!t]
    \centering
    \includegraphics[width=1.0\textwidth]{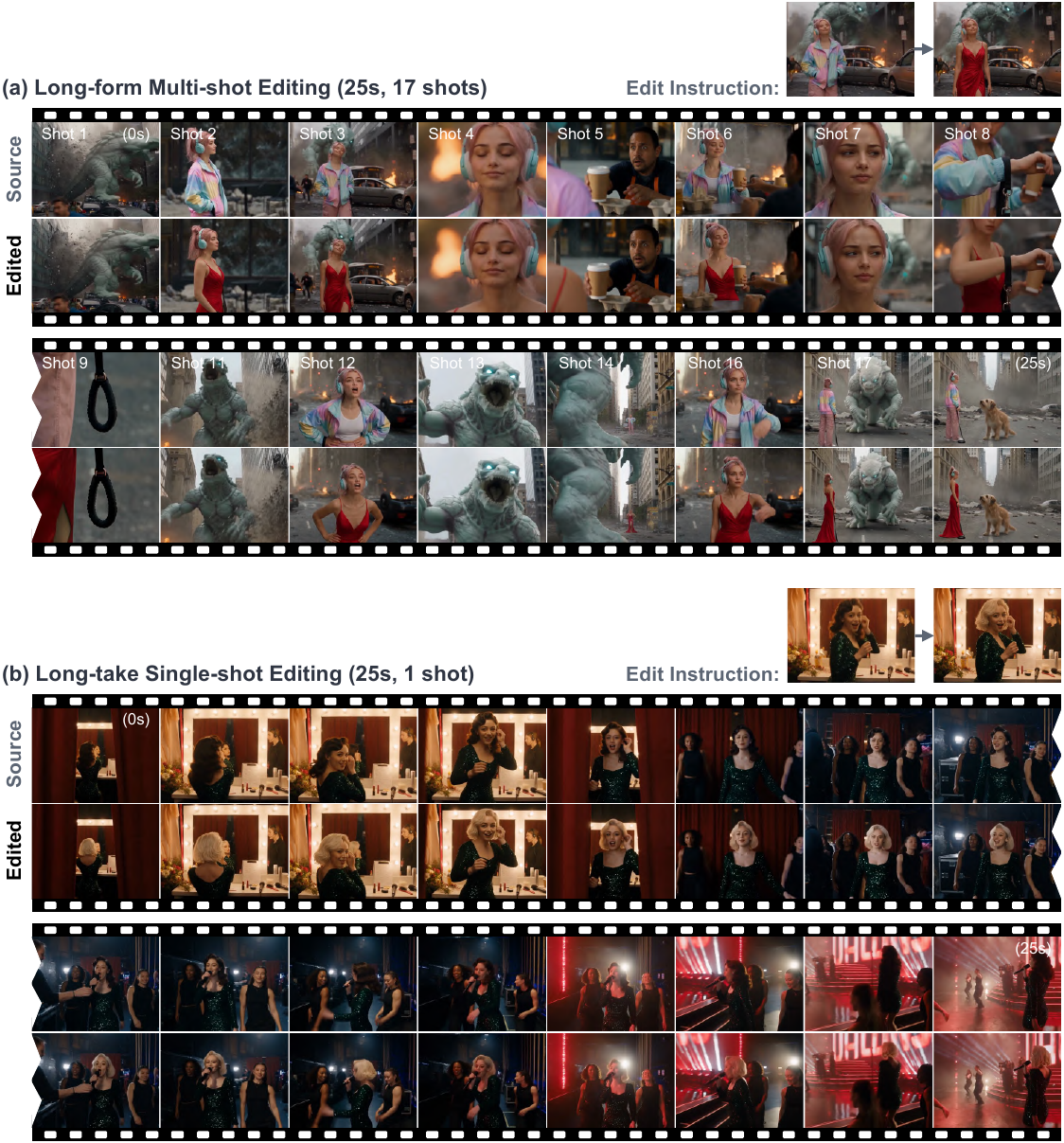}
    \caption{
    \textbf{Beyond-training-horizon editing.}
    Qualitative results on 25-second videos: (a) a 17-shot multi-shot sequence and (b) a continuous single-shot sequence. Both exceed the 15-second training horizon, demonstrating generalization to longer multi-shot and long-take videos.
    }
    \label{fig:supp-very-long}
\end{figure}

We further investigate the ability of MoviEdit to generalize beyond its training horizon. Although trained on videos of up to 15 seconds, MoviEdit can extend to substantially longer sequences with diverse shot structures. Fig.~\ref{fig:supp-very-long} presents qualitative results on two 25-second videos: (a) a multi-shot video comprising 17 shots and (b) a continuous single-shot video. The former demonstrates consistent editing across frequent shot transitions, substantially exceeding both the duration and shot count seen during training, while the latter demonstrates sustained editing consistency over an extended continuous sequence. These results illustrate the ability of MoviEdit to handle longer editing horizons across different temporal structures.

\end{document}

%% file: math_commands.tex
\usepackage{amsmath,amsfonts,bm}

\def\eqref#1{equation~\ref{#1}}

\def\1{\bm{1}}

\DeclareMathAlphabet{\mathsfit}{\encodingdefault}{\sfdefault}{m}{sl}
\SetMathAlphabet{\mathsfit}{bold}{\encodingdefault}{\sfdefault}{bx}{n}